\documentclass[final,3p,times,twocolumn]{elsarticle}

\usepackage{amssymb}
\usepackage{amsmath}
\usepackage{optidef}
\usepackage{subcaption}
\usepackage{times}
\usepackage{enumerate}
\usepackage{latexsym}
\usepackage{pdfpages}
\usepackage{adjustbox}
\usepackage{enumitem}
\usepackage{subcaption}
\usepackage{graphicx}
\usepackage{xcolor}
\usepackage{amsmath}
\usepackage{textcomp}
\usepackage{array, multirow, bigdelim, makecell, booktabs} 
\usepackage{url}
\usepackage{hyperref}
\newcommand{\blue}[1]{\textcolor{blue}{#1}}
\usepackage{setspace}
\usepackage[lined,boxed,commentsnumbered,procnumbered,ruled]{algorithm2e}
\usepackage[T1]{fontenc}
\usepackage{titlesec}
\makeatletter
\def\@xfootnote[#1]{%
  \protected@xdef\@thefnmark{#1}%
  \@footnotemark\@footnotetext}
\makeatother
 \SetKwFunction{FRecurs}{FnRecursive}%
\usepackage[utf8]{inputenc}
\usepackage{algpseudocode}  
\usepackage{microtype}

\journal{Arxiv}

\begin{document}

\begin{frontmatter}



\title{ATSumm: Auxiliary information enhanced approach for abstractive disaster Tweet Summarization with sparse training data}

\author[label1]{Piyush Kumar Garg}
\affiliation[label1]{organization={Computer Science \& Engineering Department, Indian Institute of Technology Patna},
            state={Bihar},
            country={India}}
\author[label2]{Roshni Chakraborty}
\author[label1]{Sourav Kumar Dandapat}

\affiliation[label2]{organization={Institute of Computer Science, University of Tartu},
            country={Estonia}}

\begin{abstract}
The abundance of situational information on Twitter poses a challenge for users to manually discern vital and relevant information during disasters. A concise and human-interpretable overview of this information helps decision-makers in implementing efficient and quick disaster response. Existing abstractive summarization approaches can be categorized as sentence-based or key-phrase-based approaches. This paper focuses on sentence-based approach, which is typically implemented as a dual-phase procedure in literature. The initial phase, known as the extractive phase, involves identifying the most relevant tweets. The subsequent phase, referred to as the abstractive phase, entails generating a more human-interpretable summary. In this study, we adopt the methodology from prior research for the extractive phase. For the abstractive phase of summarization, most existing approaches employ deep learning-based frameworks, which can either be pre-trained or require training from scratch. However, to achieve the appropriate level of performance, it is imperative to have substantial training data for both methods, which is not readily available. This work presents an Abstractive Tweet Summarizer (ATSumm) that effectively addresses the issue of data sparsity by using auxiliary information. We introduced the Auxiliary Pointer Generator Network (AuxPGN) model, which utilizes a unique attention mechanism called \textit{Key-phrase attention}. This attention mechanism incorporates auxiliary information in the form of key-phrases and their corresponding importance scores from the input tweets. We evaluate the proposed approach by comparing it with $10$ state-of-the-art approaches across $13$ disaster datasets. The evaluation results indicate that ATSumm achieves superior performance compared to state-of-the-art approaches, with improvement of $4-80\%$ in ROUGE-N F1-score.

\end{abstract}




\end{frontmatter}

\section{Introduction} \label{s:intro}
\par During a disaster, emergency responders, government agencies, and non-governmental organizations (NGOs) require event information to ensure quick and effective support. Social media platforms, like Twitter, play an important role in providing information during disaster events. However, there are several challenges in identifying relevant information from these tweets manually, such as the huge number of tweets and the informal language. Therefore, existing research works~\cite{dutta2019community, sharma2019going, rudra2018identifying, garg2023ontodsumm} have proposed disaster tweet summarization approaches to automatically process these huge number of tweets to generate a concise summary.


\par Existing disaster tweet summarization approaches are either extractive~\cite{garg2023ontodsumm, Garg2022Entropy, rudra2018identifying, dutta2018ensemble} or abstractive approaches~\cite{faghihi2022crisisltlsum, vitiugin2022cross, lin2021preserve, rudra2019summarizing}. Extractive approaches aim to identify the most relevant tweets from a set of input tweets, ensuring comprehensive information coverage while minimizing redundancy. On the other hand, abstractive summarization approaches strive to generate a more human-interpretable summary while still maintaining comprehensive information coverage and minimizing redundancy. Existing abstractive summarization approaches on the disaster can be categorized based on the presentation of the final summary into two categories: 1) Sentence-based presentation~\cite{vitiugin2022cross, lin2021preserve, faghihi2022crisisltlsum, rudra2019summarizing, rudrapal2019new} where the final summary consists of a number of full-length sentences; and 2) Key-phrase-based presentation~\cite{nguyen2022rational, nguyen2022towards} where the final summary consists of key-phrases. In this paper, we focus on generating a sentence-based abstractive summary.

\begin{table*}[!ht]
    \caption{Table shows F1-score of ROUGE-1, 2, and L of the summaries generated by various baselines on CNN/Daily Mail dataset. [Note: ``AuxInfo'' represents the auxiliary information of tweets in terms of tweets key-phrases and their importance scores.]}
    \label{table:results3}
    \centering 
    \resizebox{0.6\linewidth}{!}{\begin{tabular}{lccc}
        \hline
        \textbf{Approach} & \textbf{ROUGE-1}    & \textbf{ROUGE-2} & \textbf{ROUGE-L} \\ \cline{2-4}
                          & \textbf{F1-score}   &\textbf{F1-score} & \textbf{F1-score} \\ \hline
         
        PGN~\cite{see2017get} + AuxInfo (100\% training data)  & 0.39 & 0.19 & 0.42 \\
        PGN~\cite{see2017get} + AuxInfo (50\% training  data)   & 0.38 & 0.18 & 0.41 \\
        PGN~\cite{see2017get} + AuxInfo (25\% training data)   & 0.39 & 0.18 & 0.41 \\
        PGN~\cite{see2017get} + AuxInfo (10\% training data)   & 0.38 & 0.17 & 0.40 \\
        PGN~\cite{see2017get} + AuxInfo (1\% training data)    & 0.37 & 0.17 & 0.40 \\ \hline
        PGN~\cite{see2017get} (100\% training data)              & 0.31 & 0.10 & 0.34 \\
        PGN~\cite{see2017get} (50\% training data)               & 0.31 & 0.11 & 0.35 \\
        PGN~\cite{see2017get} (25\% training data)               & 0.30 & 0.09 & 0.36 \\
        PGN~\cite{see2017get} (10\% training data)               & 0.28 & 0.08 & 0.32 \\
        PGN~\cite{see2017get} (1\% training data)                & 0.27 & 0.07 & 0.30 \\ \hline
        PGN~\cite{see2017get}                               & 0.31 & 0.10 & 0.34 \\
        PTESumm~\cite{rudrapal2019new}                      & 0.28 & 0.08 & 0.32 \\  
        PEGASUS~\cite{zhang2020pegasus}                     & 0.28 & 0.08 & 0.29 \\  
        Longformer~\cite{beltagy2020longformer}             & 0.35 & 0.14 & 0.39 \\  
        ProphetNet~\cite{qi2020prophetnet}                  & 0.25 & 0.09 & 0.27 \\  
        T5~\cite{raffel2020exploring}                 & 0.36 & 0.15 & 0.38 \\  
        BART~\cite{lewis2020bart}                           & 0.37 & 0.16 & 0.40 \\  \hline
    \end{tabular}}
\end{table*}

\par Existing sentence-based abstractive disaster tweet summarization approaches~\cite{vitiugin2022cross, lin2021preserve, faghihi2022crisisltlsum, rudra2019summarizing, rudrapal2019new} generate the final summary in two subsequent phases: 1) Phase-I, which is an extractive phase; and 2) Phase-II, which is an abstractive phase. 
Existing unsupervised approaches~\cite{vitiugin2022cross, lin2021preserve, faghihi2022crisisltlsum} used for Phase-I could not cluster tweets properly due to the overlapping keywords between clusters, while the existing supervised approach~\cite{rudra2019summarizing} used for Phase-I requires a huge amount of labelled data, which incurs cost as well as time. 
However, a supervised method by Garg et al.~\cite{garg2024ikdsumm} handles the sparsity of the training data by augmenting auxiliary information in terms of key-phrases of tweets. 
So for the extractive phase of the proposed approach, we directly adopt the methodology proposed by Garg et al.~\cite{garg2024ikdsumm}. For Phase-II, i.e., abstractive-phase, in literature, there are either graph-based approach~\cite{rudra2019summarizing} or deep learning-based approaches~\cite{raffel2020exploring, see2017get}. For the graph-based approach, we find the work proposed by Rudra et al.~\cite{rudra2019summarizing}. The advantages of the graph-based approach is that it does not depend on the availability of training dataset and effectively handles disaster-specific information. For example, Rudra et al.~\cite{rudra2019summarizing} generate the summary sentences by traversing the constructed word graph, where nodes are bigrams (adjacent words), and edges are bigrams relationships between two nodes. However, this approach does not consider semantic similarity among words while creating the word graph, which can inherently lead to redundancy in the summary. On the other hand, existing deep-learning-based approaches can be further categorised into: 1) Pre-trained model-based approaches and 2) Approaches where the models are trained from scratch. The problem with using a pre-trained model for abstractive summarization is that the model is trained with a very generic corpus, which fails to handle domain-specific challenges for summary creation. For example, Vitiugin et al.~\cite{vitiugin2022cross} utilizes a pre-trained T5 model, which fails to ensure high performance as they can not handle disaster-specific challenges. On the other hand, the only methodology found in literature used for abstractive summarization by training the model from scratch is the sequence-to-sequence (Seq2Seq) model. To train an existing Seq2Seq model from scratch requires a good amount of training data, which is not readily available. 
To handle the unavailability of the disaster-specific training data, Lin et al.~\cite{lin2021preserve} trains a Seq2Seq model, i.e., PGN~\cite{see2017get} model using available CNN/Daily Mail news dataset. However, due to the training on non-disaster-specific data, it does not give the desired performance in result. 
Therefore, our main aim of this work is to modify the existing PGN architecture so that it can handle the sparse data problem by utilizing the domain knowledge.

\par Deep learning-based architecture learns the feature weight through the repeated feedback mechanism. 
Given a sufficient size of training data, a deep learning-based framework can automatically identify and learn the importance of features. However, in the absence of sufficient training data, it may be difficult for the model to identify and learn the features importance. 
Therefore, in this paper, we try to address the above limitation by computing features importance (key-phrase and its importance) separately using ontology and providing it as the input to the deep learning model. There exists deep-learning-based architecture, which uses this kind of hybrid model where auxiliary information is provided by handcrafted features~\cite{wen2022sememe, ni2022two}, we are adopting the same to solve the sparse data problem in this paper. 
The effectiveness of auxiliary information is verified using a small experiment, where we use a fraction of CNN/Daily Mail dataset~\cite{hermann2015teaching} along with auxiliary information to summarize documents related to a news domain. As shown in Table~\ref{table:results3}, experiment results clearly indicate that it can produce a comparable summary even when only $1$\% of training data is used with auxiliary information.  


\par In this paper, we propose a Two-Phase approach, \textbf{ATSumm} (\textbf{A}bstractive \textbf{T}weet \textbf{Summ}arizer), specifically designed to create a summary of a disaster event, where in Phase-I, we follow Garg et al.~\cite{garg2024ikdsumm} to identify the most relevant tweets automatically by effectively utilizing augmenting auxiliary information as domain knowledge from an existing disaster ontology. For Phase-II, we propose an extended version of the PGN model, which we refer to as a \textbf{AuxPGN} (\textbf{A}uxiliary \textbf{P}ointer \textbf{G}enerator \textbf{N}etwork) model that utilizes auxiliary information in the form of key-phrases from the extracted tweets along with their importance score to ensure the AuxPGN model effectively work with the sparse training data. The novelty of AuxPGN is that it utilizes a novel attention mechanism referred to as \textit{\textbf{Key-phrase attention}} guided by source tweets as well as its key-phrases and their importance scores as auxiliary information generated by DRAKE~\cite{garg2024ikdsumm}. The key-phrase generated by DRAKE consists of the most relevant phrase (key-phrase) related to a disaster event. 
ATSumm can ensure better performance compared to existing state-of-the-art abstractive summarization approaches by $4-61$\%, $7-80$\%, and $4-64$\%, in terms of ROUGE-1, 2, and L, respectively. We also perform ablation experiments where we analyze the performance of each phase of ATSumm with $10$ different variants of ATSumm to justify the model of each phase in ATSumm. We provide a human evaluation of the effectiveness of ATSumm in the abstractive summary generation based on five different quality metrics, i.e., \textit{Fluency}, \textit{Readability}, \textit{Conciseness}, \textit{Relevance}, and \textit{Non-redundancy}. Finally, through experimental analysis, we show that ATSumm outperforms the five existing transformer models fine-tuned by $3.92-33.33$\% in terms of ROUGE-N F1-score on the ARIES dataset, which we discussed in the Subsection~\ref{s:finetune}. 

\par Our major contributions can be summarized as follows: 
\begin{enumerate} 
    \item To the best of our knowledge, our proposed \textbf{ATSumm} is the first model on abstractive disaster tweet summarization, which utilizes auxiliary information in a deep-learning framework to handle the problem of sparse training data.
    
    
    
    \item We propose a novel attention mechanism, \textit{\textbf{Key-phrase attention}} that incorporates the tweet's key-phrases and their importance scores as auxiliary information into the proposed AuxPGN model.
    
    \item We prepare a new \textbf{ARIES} (\textbf{A}bst\textbf{R}active d\textbf{I}saster tw\textbf{E}et \textbf{S}umm-arization) dataset. To the best of our knowledge, ARIES is the first disaster domain training dataset for the AuxPGN/PGN model, containing $7500$ training samples where each training sample comprises $400$ input words and corresponding $200$ words output summary. 
    
    \item Additionally, we prepare and release the ground-truth summaries of $13$ different disaster events belonging to the different types and locations (used for evaluation). 
    
    \item Evaluation results clearly indicate that the proposed approach, ATSumm, outperforms existing state-of-the-art abstractive summarization approaches for all $13$ disaster events.
\end{enumerate}

\par Rest of the paper is organized as follows. In Section~\ref{s:rworks}, we discuss the existing research works of abstractive tweet summarization. Section~\ref{s:data} discusses the ARIES training dataset and evaluation datasets in detail. In Section~\ref{s:prop}, we discuss our proposed approach in detail, followed by experiments and comparison results in Section~\ref{s:expt}. We discuss the ablation experiments in Subsection~\ref{s:ablexp}. Finally, we conclude our paper and provide future works in Section~\ref{s:con}.

\section{Related Works} \label{s:rworks}
\par Summarization refers to creating a short and concise overview of an input text while preserving all its important key information~\cite{liu2024automatic, qu2022knowledge, srivastava2022topic, zhao2022mrs, xiao2022fusionsum}. Several researchers have proposed automatic text summarization approaches for various domains, including legal documents summarization~\cite{jain2023bayesian, bhattacharya2021incorporating}, news documents summarization~\cite{ahuja2022aspectnews, hernandez2022language, curiel2021online}, scientific documents summarization~\cite{saini2023multi, mishra2022scientific}, tweets summarization~\cite{chakraborty2019tweet, chakraborty2017network}, product reviews summarization~\cite{boorugu2020survey, komwad2022survey} etc. Social media platforms, such as Twitter, have gained prominence as a means for users to share personal updates, opinions, and information about a real-time event. For real-time updates, Twitter is considered one of the best social media platforms~\cite{priya2019should}. However, the huge amount of information posted on Twitter makes it very difficult for any human being to get relevant information. Therefore, the research community devoted a lot of attention to automatic tweet summarization. 

\par While all tweet summarization approaches share basic objectives, such as maximizing coverage and decreasing redundancy, the specific knowledge required may vary depending on the domain of the summarizer. Domain expertise enables the summarizer to comprehend the significance of a certain keyword or phrase. The research community proposed automatic summarization approaches for different domains, such as 
sports game tweets summarization~\cite{goyal2019multilevel, huang2018event, gillani2017post}, news tweets summarization~\cite{zheng2021tweet, duan2019across, chakraborty2017network}, disaster tweet summarization~\cite{garg2023ontodsumm, garg2023portrait, garg2024ikdsumm}, political event tweets summarization~\cite{panchendrarajan2021emotion, kim2014tweet}, etc. Existing disaster tweet summarization approaches can be classified into extractive and abstractive. Existing abstractive summarization approaches on the disaster can be further categorized based on the generation of the final summary into two classes: 1) Sentence-based~\cite{vitiugin2022cross, lin2021preserve, faghihi2022crisisltlsum, rudra2019summarizing, rudrapal2019new} where the final summary consists of a number of full-length sentences; and 2) Key-phrase-based~\cite{nguyen2022rational, nguyen2022towards} where the final summary consists of key-phrases. In this work, we only focus on an abstractive sentence-based approach, and therefore, we discuss only the related work of abstractive sentence-based disaster tweet summarization approaches in detail. Existing approaches for abstractive sentence-based disaster tweet summarization are either unsupervised or supervised. We discuss each of them next.

\subsection{Unsupervised sentence-based approaches}
\par Existing unsupervised abstractive sentence-based tweet summarization approaches are graph-based~\cite{rudra2019summarizing}.  
These approaches initially construct a word graph in which nodes are words and edges are between consecutive words and then traverse the word graph to generate new sentences. In the final summary, it selects the generated new sentences that provide the most information coverage and diversity. For example, Rudra et al.~\cite{rudra2019summarizing} propose an extractive-abstractive tweet summarization framework for disaster events where they initially identify the set of important tweets based on the importance of their content words (i.e., nouns, verbs, adjectives, and numerals). Further, from important tweets, they created a word-graph where nodes are bigrams (adjacent words), and edges are between consecutive words. Finally, they generate sentences by word-graph traversal and create a summary by selecting those sentences into the summary, which provides maximum information coverage and diversity in the final summary using the Integer Linear Programming (ILP) objective function. 
However, this approach does not consider semantical similarity among words while creating the word graph, which can inherently lead to redundancy in the summary.

\subsection{Supervised sentence-based approaches}
\par Recent abstractive supervised tweet summarization approaches are based on sequence-to-sequence problems. These abstractive models/approaches are classified into two categories based on their training strategy: 1) model training from scratch and 2) pre-trained model. 
We next discuss each of these categories in detail.

\subsubsection{Abstractive models used as a pre-trained model}\label{s:trainpre}

\par The introduction of transformer architectures has revolutionized the field of automatic document summarization, as they have been shown to be highly effective in various natural language Seq2Seq tasks. Microsoft's Unified Pretrained Language Model (UniLM)~\cite{dong2019unified} stands out as one of the leading Transformer-based models. UniLM, designed by Microsoft, adopts a model architecture similar to BERT (large) \cite{kenton2019bert} and is pre-trained using English Wikipedia and BookCorpus~\cite{zhu2015aligning}. Another important model is ProphetNet~\cite{qi2020prophetnet}, which addresses the traditional Seq2Seq optimization problem by predicting the n-next tokens instead of a single token at a time. The next model is Pegasus~\cite{zhang2020pegasus}, which uses a novel summarization-specific pre-training objective called Gap Sentence Generation (GSG) that involves masking out important sentences from input sentences and then generating those sentences from the remaining sentences. The next model is the Text-to-Text Transfer Transformer (T5) model introduced by Google~\cite{raffel2020exploring}, which utilizes transfer learning on the basic transformer architecture proposed by Vaswani et al.~\cite{vaswani2017attention}. Further, the Bidirectional Auto-Regressive Transformers (BART) model introduced by Facebook~\cite{lewis2020bart} utilizes a denoising autoencoder that is trained on corrupted text with an arbitrary noising function to learn the reconstruction of the input text. The final model is Longformer-Encoder-Decoder (LED)~\cite{beltagy2020longformer}, which utilizes a local windowed attention mechanism to only attend to a small window of tokens at a time and compute the attention weights.

\par Existing abstractive tweet summarization work~\cite{vitiugin2022cross} follows a two-phase architecture to generate a meaningful summary, where the first phase is extractive, and the second is abstractive. In the extractive phase, it initially identifies different clusters using a K-means clustering-based approach and then selects the representative tweets from each cluster. In the abstractive phase, it utilizes diversification operations to create a human-interpretable summary of the tweets related to a disaster event by using a pre-trained T5 model. However, the pre-trained model is trained with a very generic corpus, which fails to handle domain-specific challenges for summary creation. 

\subsubsection{Abstractive models training from scratch}\label{s:trainsc}
\par Nallapati et al.~\cite{nallapati2016abstractive} proposed an Attention-based Encoder-Decoder model (AED), one of the first deep learning architectures to demonstrate efficacy in abstractive summarization. The attention mechanism relies upon the concept that only specific input portions are relevant during each generation step. Therefore, to take advantage of the AED architecture, See et al.~\cite{see2017get} propose a Pointer Generator Network (PGN) model, which solves the Out-Of-Vocabulary (OOV) word problem by allowing the summarizer to copy words from the input text and generate new text.

\par Recent existing sentence-based abstractive tweet summarization approaches~\cite{lin2021preserve, rudrapal2019new} follow a two-phase architecture to create a more human interpretable summary where the first phase is extractive, and the second is abstractive. In the extractive phase, they intend to extract a set of important tweets from the input tweets using either a clustering-based approach~\cite{lin2021preserve} or a ranking-based approach~\cite{rudrapal2019new} that ensures maximum information coverage without redundancy. However, the clustering-based approach could not cluster tweets properly due to the overlapping keywords between clusters, while the ranking method is very generic, which fails to compute the proper importance of a tweet with respect to a disaster. In the abstractive phase, they take the output of the extractive phase as an input and then create a concise and more human-interpretable summary of a predefined length using the existing PGN model. They utilize the available CNN/Daily Mail dataset~\cite{hermann2015teaching, nallapati2016abstractive} to train the PGN model. However, the PGN model requires huge amounts of training data, which is difficult to obtain in disaster domains where we have sparse data for training. Therefore, to address this limitation, we proposed an extended/modified version of the PGN model as Auxiliary Pointer Generator Network (AuxPGN) model that utilizes the auxiliary information (i.e., key-phrases and their importance scores) extracted from the input text and provides this auxiliary information through novel key-phrase attention with input text while training the AuxPGN model. As a result, the AuxPGN model does not require extensive training data and performs better with various start-of-the-art approaches for abstractive summary generation.

\section{Datasets} \label{s:data} 
\par In this Section, we initially discuss the \textbf{A}bst\textbf{R}active d\textbf{I}saster tw\textbf{E}et \textbf{S}ummarization (ARIES) dataset used for training. Then we discuss $13$ evaluation datasets belonging to different disaster types and locations.

\subsection{Training Dataset Details} \label{s:traindata}        

\par Although the existing research works~\cite{rudra2019summarizing, vitiugin2022cross} have provided abstractive ground-truth summaries for six disaster events. All these works consider all the tweets for a disaster event as input tweets, and the corresponding summary consists of $250$ words. However, the PGN/AuxPGN model expects an input of size $400$ words and produces a summary output of $200$ words. Hence, we can not use the existing datasets with their ground-truth summaries for training in our proposed AuxPGN model. Therefore, in this paper, we manually create a training dataset to train the AuxPGN model. To prepare a training dataset, we found $30$ available disaster datasets~\cite{alam2021humaid, olteanu2014crisislex, priya2020taqe}, which do not have any available ground-truth summaries. For each disaster dataset, we follow the method proposed by Nallapati et al.~\cite{nallapati2016abstractive} to create a training dataset. We use all the tweets of a dataset as input and generate several chunks (samples) of $400$ words. Finally, we got a total of $7500$ samples from all the $30$ disaster events. We ask two human annotators to write a summary of each sample manually based on their understanding and wisdom. We select these annotators through the Quality Assessment Evaluation proposed by Garg et al.~\cite{garg2023portrait}. The annotators are undergraduate students between the ages of $20-30$ who are fluent in English, are regular Twitter\footnote{https://twitter.com} users, and are not part of this project. To ensure a high-quality ground-truth summary, we follow a two-step annotation procedure, wherein in the first step, an annotator goes through all the given words of a sample and then writes a summary of approximately but not more than $200$ words in length that includes all the relevant information related to that event with minimum repetition. Then, in the second step, another annotator who did not participate in the first step reads all the given words of a sample and the first annotator's annotated summary and then corrects the already annotated summary to create a final summary if it contains irrelevant or incorrect information.

\begin{table*}
    \caption{Table shows the details of $D_1$-$D_{13}$ datasets.}
    \label{table:datasets}
    \centering
    \resizebox{0.65\linewidth}{!}{\begin{tabular} {clccccc}
        \hline
        {\bf Num} & {\bf Dataset Name} & {\bf Year} & {\bf \# Tweets} & {\bf Continent} & {\bf Disaster type}\\\hline  
        
        $D_1$ & \textit{Sandy Hook Elementary School Shooting}      & 2012 & 2080 & USA  & Man-made\\ 
        $D_2$ & \textit{Uttrakhand Flood}                           & 2013 & 2069 & Asia & Natural \\ 
        $D_3$ & \textit{Hagupit Typhoon}                            & 2014 & 1461 & Asia & Natural \\ 
        $D_4$ & \textit{Hyderabad Blast}                            & 2013 & 1413 & Asia & Man-made\\ 
        $D_5$ & \textit{Harda Twin Train Derailment}                & 2015 & 4171 & Asia & Man-made\\ 
        $D_6$ & \textit{Los Angeles Airport Shooting}               & 2013 & 1409 & USA & Man-made \\ 
        $D_7$ & \textit{Matthew Hurricane}                          & 2016 & 1654 & USA & Natural\\ 
        $D_8$ & \textit{Puebla Mexico Earthquake}                   & 2017 & 2015 & USA & Natural \\ 
        $D_9$ & \textit{Pakistan Earthquake}                        & 2019 & 1958 & Asia & Natural\\ 
        $D_{10}$ & \textit{Midwestern U.S. Floods}                  & 2019 & 1880 & USA & Natural\\ 
        $D_{11}$ & \textit{Kaikoura Earthquake}                     & 2016 & 2195 & Oceania & Natural \\ 
        $D_{12}$ & \textit{Cyclone Pam}                             & 2015 & 1508 & Oceania & Natural\\ 
        $D_{13}$ & \textit{Greece Wildfire}                         & 2018 & 1282 & Europe & Natural\\  \hline
    \end{tabular}}
\end{table*}

\subsection{Evaluation Dataset and Pre-processing:}
\par We evaluate the performance of ATSumm on $13$ different disaster datasets, which as are follows:

\begin{enumerate}
    \item \textit{$D_1$}: This dataset is prepared by Dutta et al.~\cite{dutta2018ensemble} on the basis of the \textit{Sandy Hook Elementary School Shooting}\footnote{https://en.wikipedia.org/wiki/Sandy\_Hook\_Elementary\_School\_shooting} on December, $2012$ in which around $20$ people were killed by a $20$-year-old shooter. 
    
    \item \textit{$D_2$}: This dataset is also prepared by~\cite{dutta2018ensemble} on the basis of the dreadful floods from the \textit{Uttarakhand Flood}\footnote{https://en.wikipedia.org/wiki/2013\_North\_India\_floods} on June, $2013$ in which $5,700$ people were died and around $128$ houses were damaged.

    \item \textit{$D_3$}: This dataset is also prepared by~\cite{dutta2018ensemble} on the basis of the \textit{Hagupit Typhoon}\footnote{https://en.wikipedia.org/wiki/Typhoon\_Hagupit\_(2014)} on December, $2014$ in which $18$ people were died and caused an estimated damage were \$$114$ million USD.
    
    \item \textit{$D_4$}: This dataset is also prepared by~\cite{dutta2018ensemble} on the basis of the two consecutive bomb blast from the \textit{Hyderabad Blast}\footnote{https://en.wikipedia.org/wiki/2013\_Hyderabad\_blasts} on February, $2013$ in which $18$ people were killed and more than $131$ people were injured.
    
    \item \textit{$D_5$}: This dataset is prepared by~\cite{Rudra2018extracting} on the basis of a massive train incident from the \textit{Harda Twin Train Derailment}\footnote{https://en.wikipedia.org/wiki/Harda\_twin\_train\_derailment} on August, $2015$ in which $31$ people were died and more than $100$ people were injured.
    
    \item \textit{$D_6$}: This dataset is prepared by~\cite{olteanu2015expect} on the basis of a terrorist attack from the \textit{Los Angeles Airport Shooting}\footnote{https://en.wikipedia.org/wiki/2013\_Los\_Angeles\_International\\\_Airport\_shooting} on November $2013$, in which one person was killed and more than $15$ people were injured.
    
    \item \textit{$D_7$}: This dataset is prepared by~\cite{alam2021humaid} on the basis of a devastating impact of a \textit{Hurricane Matthew}\footnote{https://en.wikipedia.org/wiki/Hurricane\_Matthew} on October, $2016$ in which $603$ people were died and caused an estimated damage were \$$2.8$ billion USD. 
    
    \item \textit{$D_8$}: This dataset is also prepared by~\cite{alam2021humaid} on the basis of a massive earthquake from the \textit{Puebla Mexico Earthquake}\footnote{https://en.wikipedia.org/wiki/2017\_Puebla\_earthquake} on September, $2017$ in which $370$ people was died and more than $6000$ people were injured.
    
    \item \textit{$D_9$}: This dataset is also prepared by~\cite{alam2021humaid} on the basis of a massive earthquake from the \textit{Pakistan Earthquake}\footnote{https://en.wikipedia.org/wiki/2019\_Kashmir\_earthquake} on September, $2019$ in which $40$ people were died and around $319$ houses were damaged.
    
    \item \textit{$D_{10}$}: This dataset is also prepared by~\cite{alam2021humaid} on the basis of a dreadful floods from the \textit{Midwestern U.S. Floods}\footnote{https://en.wikipedia.org/wiki/2019\_Midwestern\_U.S.\_floods} in which around $14$ million people were affected and caused approximately damage were \$$2.9$ billion USD.
    
    \item \textit{$D_{11}$}: This dataset is also prepared by~\cite{alam2021humaid} on the basis of a massive earthquake from the \textit{Kaikoura Earthquake}\footnote{https://en.wikipedia.org/wiki/2016\_Kaikoura\_earthquake} on November, $2016$ in which $2$ people were died and more than $57$ people were injured.
    
    \item \textit{$D_{12}$}: This dataset is prepared by~\cite{imran2016twitter} on the basis of a devastating impact of the \textit{Cyclone Pam}\footnote{https://en.wikipedia.org/wiki/Cyclone\_Pam} on March, $2015$ in which $15$ people were died and around $3300$ people were displaced.

     \item \textit{$D_{13}$}: This dataset is also prepared by~\cite{alam2021humaid} on the basis of a mass-fire from the \textit{Greece Wildfire}\footnote{https://en.wikipedia.org/wiki/2018\_Attica\_wildfires} on July, $2018$ in which $4$ people were died and around 4000 homes/buildings were damaged.
\end{enumerate}

\begin{table*}[!ht]
    \centering 
    \caption{Table lists key notations and their descriptions used in ATSumm.}
    \label{table:notation}
    \resizebox{0.6\linewidth}{!}{\begin{tabular}{|c|c|}
        \hline
        {\bf Notation} & {\bf Description}  \\ \hline 
    
        $D$             & Disaster event dataset \\\hline
        $T$             & A set of tweets in $D$ \\\hline
        $m$             & Desired length summary (i.e., $200$ words) \\\hline
        $T^{'}$         & Set of tweets extracted from Phase-I \\\hline 
        $N$             & Number of the words in $T^{'}$ \\\hline
        $h_i$           & Encoder hidden state \\ \hline
        $x_t$           & Decoder input \\ \hline
        $s_t$           & Decoder state \\ \hline
        $V$             & Fixed length vocabulary \\ \hline
        $KP_i$          & Key-phrase of $i^{th}$ indexed tweet in $T^{'}$  \\ \hline 
        $a^t$           & Key-phrase attention distribution \\ \hline
        $\gamma^t$      & Key-phrase word vector \\ \hline
        $e^t$           & Precursor attention vector \\ \hline
        $W_i$           & Word probability vector of $KP_i$ \\ \hline
        $h^{*}_{t}$     & A context vector  \\ \hline
        ${P_V}$         & Output word probability distribution of all $W \in V$ \\ \hline
        $P_{gen}$       & Generation probability \\ \hline
        $c^t$           & A coverage vector \\ \hline
        $L_t$           & A loss function \\ \hline
        $DRAKE_{Score}(KP_i)/T$ & DRAKE score of $i^{th}$ key-phrase in $T$ \\ \hline
    \end{tabular}}
\end{table*}

\textit{Dataset Pre-processing}: For pre-processing, we remove URLs, emoticons, punctuation marks, and stop words. Further, we remove hashtags and usernames as we only consider the Twitter text. We also observe that lengths less than $3$ characters words do not provide any relevant information specific to disasters~\cite{alam2018crisismmd}, so we also remove these words. We also perform lowercase conversion of the dataset text. We prepare the ground-truth summary of $200$ words in length of these $13$ datasets with the help of the two-step annotation procedure discussed in Subsection~\ref{s:traindata} and show an overview of these datasets in Table~\ref{table:datasets}.

\section{Proposed Approach}\label{s:prop}
\par In this Section, we elaborate proposed approach in detail.

\subsection{Model Overview} \label{s:Overview} 
\par
\begin{figure}[htbp]
    \centering 
    \includegraphics[width=\linewidth] {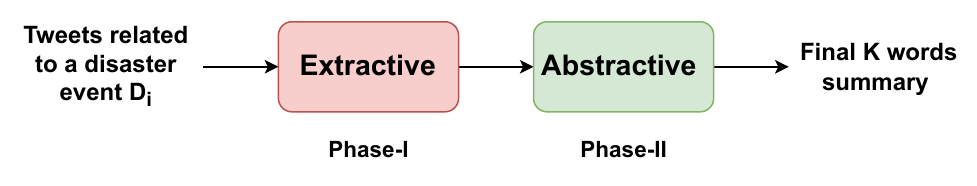}
    \caption{Figure shows an overview of ATSumm.}
    \label{figure:flowchart1}
\end{figure}

 The proposed approach, ATSumm, comprises of two phases. In Phase-I, we identify the most relevant tweets on the basis of disaster domain knowledge as auxiliary information to ensure maximum information coverage and diversity. In Phase-II, we utilize the extracted tweets from Phase-I as input to generate new sentences in a more human-interpretable manner, while maintaining the information coverage and minimizing redundancy. We propose an \textbf{A}uxiliary \textbf{P}ointer \textbf{G}enerator \textbf{N}etwork (\textbf{AuxPGN}) model that utilizes the tweets' key-phrases and their importance scores through our proposed \textit{Key-phrase attention} mechanism, and therefore, can ensure maximum disaster relevant information coverage in summary. Therefore, our proposed AuxPGN requires less data for training as it learns the most important information related to a disaster event through the importance of key-phrases. We show the overview of ATSumm in Figure~\ref{figure:flowchart1} and show all the used notations and corresponding descriptions in Table~\ref{table:notation}.  

\begin{figure*}
    \centering 
    \includegraphics[width=1.0\textwidth] {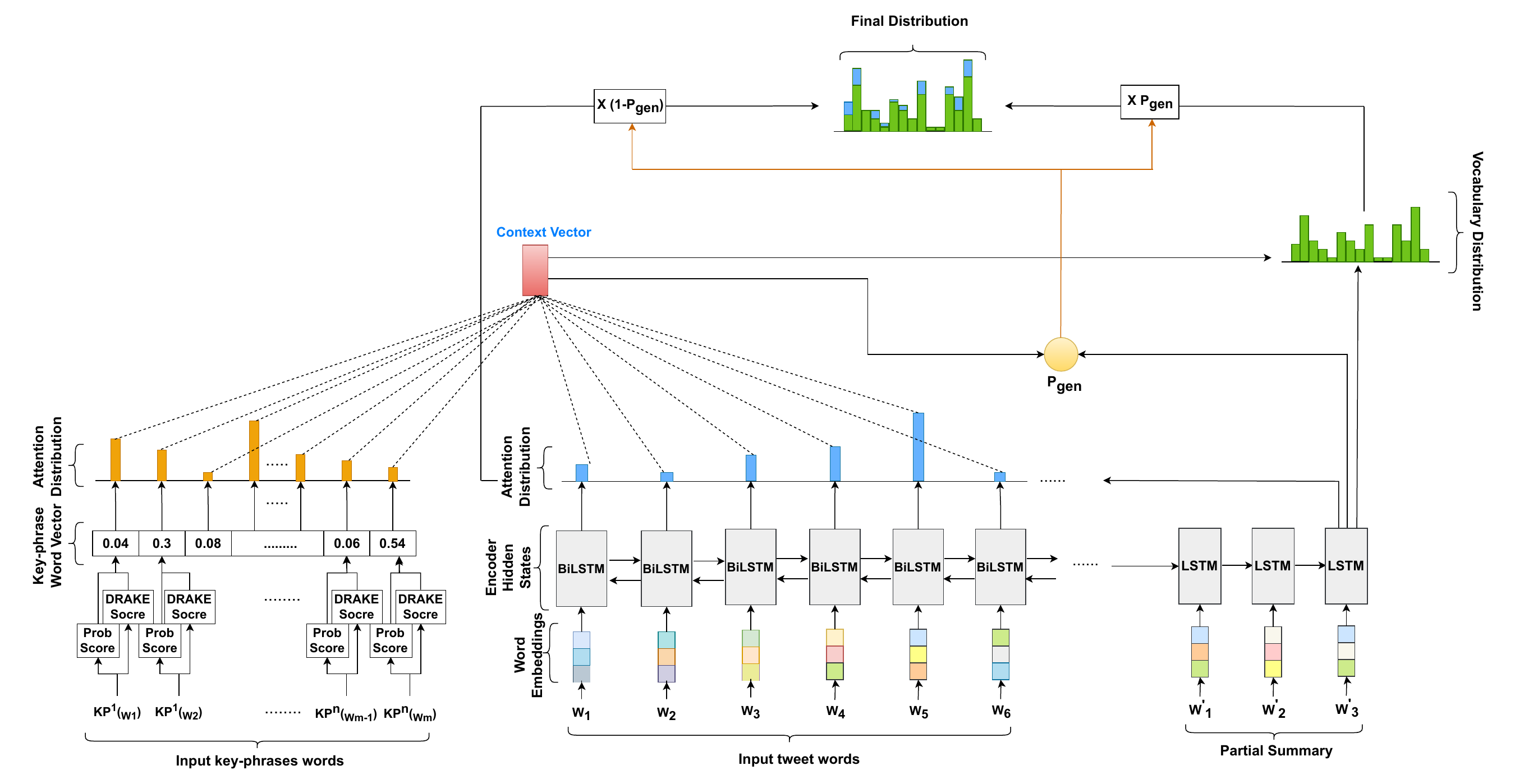}
    \caption{Figure shows the proposed architecture of AuxPGN model.}
    \label{figure:flowchart}
\end{figure*}

\subsection{Phase-I}\label{s:phase1}
\par In this phase, we follow Garg et al.~\cite{garg2024ikdsumm} to automatically identify the most relevant tweets related to a disaster event that can captures the maximum information coverage and ensures diversity in summary. Although there are several approaches available~\cite{garg2023ontodsumm, Garg2022Entropy, rudra2015extracting, dutta2018ensemble}, which identify the most relevant tweets, we follow Garg et al.~\cite{garg2024ikdsumm} for its inclusion of auxiliary information as domain-specific knowledge. Additionally, Garg et al.~\cite{garg2024ikdsumm} can ensure the best performance, i.e., $0.47-0.61$ ROUGE-1 F1-score, without any manual intervention. We got a ranked list of tweets as an output of \cite{garg2024ikdsumm} approach on the basis of their importance and diversity scores, and then we select the top-ranked tweets until the word count reached $400$~\cite{see2017get}. 
We now provide these tweets as input to Phase-II for abstractive summary generation.


\subsection{Phase-II} \label{s:phase2}
\par In Phase-II, we aim to generate new sentences as a summary, which is more human-interpretable from the tweets extracted from Phase-I. Existing abstractive tweet summarization approaches~\cite{lin2021preserve, rudrapal2019new} have proposed PGN~\cite{vinyals2015pointer} model to generate a human-interpretable summary. The PGN-based Seq2Seq model outperforms traditional Seq2Seq models for summarization by handling the out-of-vocabulary word problem. Hence, we chose the PGN-based model over other Seq2Seq models. 
However, the PGN requires a huge amount of training data, which is very difficult to obtain for disaster datasets. To handle this challenge, we propose integration of the most relevant auxiliary information with respect to each disaster dataset to PGN, namely the proposed AuxPGN Model. Therefore, we ensure significant reduction in the training dataset size. Additionally, by integration of disaster-specific most relevant auxiliary information, we can ensure a high increase in performance specific to each disaster event without incurring any human intervention. In AuxPGN, we introduce the tweet's key-phrases and their importance scores as auxiliary information generated by the DRAKE~\cite{garg2024ikdsumm}. Additionally, we introduce a novel \textit{Key-phrase attention}, which utilizes this auxiliary information into the AuxPGN, which helps reduce the training data size required for the AuxPGN model. We show an overview of the AuxPGN model in Figure~\ref{figure:flowchart} and describe the architecture next by expanding on details regarding the constituent components, \textbf{Encoder and Decoder} (Subsection~\ref{s:EncDec}), \textbf{Key-Phrase Attention} (Subsection~\ref{s:KeyAtten}), \textbf{Pointer Generator} (Subsection~\ref{s:PontGen}), \textbf{Coverage Mechanism} (Subsection~\ref{s:CoveMech}), and \textbf{Decoding} (Subsection~\ref{s:Dec}).

\subsubsection{Encoder and Decoder} \label{s:EncDec} 
\par For pre-processing, we use  Stanford CoreNLP\footnote{https://stanfordnlp.github.io/CoreNLP/} tokenizer for tokenization. We feed these tweet tokens to the encoder layer one-by-one. Although there are several encoders available, such as RNN~\cite{cho2014learning}, LSTM~\cite{greff2016lstm}, and GRU~\cite{chung2014empirical}, we choose a single-layer Bi-directional Long Short-Term Memory (BiLSTM)~\cite{graves2005framewise} model. BiLSTM is able to capture long-term dependencies and handles sequential data to generate concise and contextually relevant summaries~\cite{liu2019bidirectional}. The encoder output is a sequence of encoder hidden states $h_i$, and each hidden state is a concatenation of both the forward and backward hidden states, as in~\cite{bahdanau2014neural}. For decoder, we employ a single-layer Long Short-Term Memory (LSTM)~\cite{greff2016lstm} model, which receives the word embedding of the preceding word to decide whether to generate a word from the vocabulary or copy a word from the input sequence to create a summary.


\subsubsection{Key-Phrase Attention} \label{s:KeyAtten} 
\par We propose a novel \textit{key-phrase attention} distribution $a^t$, which is a combination of the usual attention weights as in~\cite{bahdanau2014neural} and a \textit{key-phrase word vector} derived from a key-phrase model. We follow Garg et al.~\cite{garg2024ikdsumm} for key-phrase identification as it does not require any training data. It utilizes the domain knowledge from existing ontology, Empathi~\cite{gaur2019empathi} to identify the key-phrase $KP$ of each tweet in $T^{'}$, where $T^{'}$ is a set of tweets extracted from Phase-I. We calculate a key-phrase word vector, $\gamma^t$ of $|V|$-dimension (where $V$ is a fixed length vocabulary) considering all the words in $T^{'}$ as: 

\begin{equation}
\gamma^t = \begin{cases}
             \sum\limits_{i} (DRAKE_{Score}(KP_i)/T^{'}).W_i, \quad If \: W_{i} \in KP_i \\
             0, \: elsewhere.
       \end{cases} \quad
\end{equation}

where $DRAKE_{Score}(KP_i)/T^{'}$ is the score of $i^{th}$ key-phrase in $T^{'}$ by ~\cite{garg2024ikdsumm}  and $W_i$ is a $|V|$ dimensional vector represents word probabilities of $KP_i$ calculated as $W_{i, j \in KP_i} = P(word_j/KP_i)$ in vocabulary $|V|$. For each decoding step $t$, we compute a final attention distribution, i.e., \textit{key-phrase attention distribution} $a^t \in R^N$ ($N$ is the number of words in $T^{'}$), which is the combination of both precursor attention vector, $e^t \in R^N$ and key-phrase word vector, $\gamma^t$ as:

\begin{equation}
e^t = {v^{T^{'}}}{tanh({W_h}{h_k}+{W_s}{s_t}+b_{att})}
\label{eq:kpvector}
\end{equation}

\begin{equation}
a^t = f(e^t, {\gamma}^t)
\end{equation}

where $h_k \in R^n$ represents $k^{th}$ encoder hidden state, $s_t \in R^l$ represents decoder state, and $v$, $W_h$, $W_s$, $b_{att}$ are the learnable parameters. Further, we use function $f$ to combine both the precursor attention vector and the key-phrase word vector as: 

\begin{equation}
f = w_1*softmax(e^t) + w_2*softmax(\bar{\gamma}^t)
\label{eq:function}
\end{equation}
\begin{equation}
   \textrm{s.t.} \quad  w_1 + w_2 =1 \quad and \quad w_1, w_2 \in (0, 1)
    \label{eq:Cweight1} 
\end{equation}

where $w_1$ and $w_2$ are the weights of $e^t$ and $\bar{\gamma}^t$, respectively, and we decide these weights through an experiment analysis discussed in Subsection~\ref{s:thre}. $\bar{\gamma}^t \in R^N$ is a reduced key-phrase word vector of size $N$, created by selecting the $N$ terms of ${\gamma}^t$. This attention distribution instructs the decoder about where to concentrate while creating the next word. Further, we derive a context vector, $h^{*}_{t}$, which is a weighted sum of the encoder hidden states and $a^t$ calculated as:

\begin{equation}
h^{*}_{t} = \sum\limits_{k} a^{t}_{k}.h_{k}
\end{equation}

Finally, we generate an output distribution probability ${P_V}$ over all words in $V$ by concatenating $h^{*}_{t}$ with $s_t$ and then linearly transforming it and pass it through a softmax activation function as:

\begin{equation}
P_V = softmax(U[s_t, h^{*}_{t}] + b)
\end{equation}

where $U$ and b are the learnable parameters.

\subsubsection{Pointer Generator} \label{s:PontGen}
\par The subsequent component of AuxPGN is a copy mechanism~\cite{gu2016incorporating}. The idea behind the pointer generator is to circumvent the limitations of pure abstraction when dealing with factual information that necessitates copying from the input to generate a correct summary. The basic encoder-decoder model~\cite{nallapati2016abstractive} does not capture factual information accurately while generating the summary. Therefore, Vinyals et al.~\cite{vinyals2015pointer} introduce a pointer network in the machine translation domain. We utilize the concept of pointer generators in our AuxPGN, in order to provide the choice of selection to the model whether to generate a word from a fixed vocabulary or to copy it from the input source directly when needed.  

\par We define a generation probability, $P_{gen}$, such that $P_{gen} \in [0,1]$. For time step $t$, we calculate $P_{gen}$ from the context vector $h^{*}_t$, the decoder state $s_t$ and the decoder input $x_t$ as:


\begin{equation}
P_{gen} = \sigma(w^{T}_{h^{*}} h^{*}_{t} + w^{T}_{s} s_{t} + w^{T}_{x} x_{t} + b_{p})
\end{equation}

where, the vectors $w^{T}_{h^{*}}$, $w^{T}_{s}$, $w^{T}_{x}$ and scalar $b_{p}$ are the learnable parameters and $\sigma$ is a sigmiod activation function. Further, we define `extended vocabulary' as a union of all the input words with $|V|$ and then calculate the final probability distribution over extended vocabulary as:

\begin{equation}
P(w) = P_{gen}*P_V(w) + (1-P_{gen}) * \sum\limits_{\forall i:w_{i}=w} a^{t}_{i}
\label{eq:prob}
\end{equation}

We observe from Equation~\ref{eq:prob} that if a word $w$ is an out-of-vocabulary (OOV) word, then $P_V(w)$ is $0$. Similarly, if $w$ does not exist in the input, $\sum\limits_{\forall i:w_{i}=w} a^{t}_{i}$ equals $0$. Therefore, the pointer generator model efficiently solves the problem of factual errors in the output summary by including the ability to switch between OOV words and words from $|V|$. 

\subsubsection{Coverage Mechanism} \label{s:CoveMech}
\par Existing attention-based recurrent neural networks model~\cite{nallapati2016abstractive} suffer from the text repetition problem while generating the summary. To solve this, in AuxPGN, we adopt a coverage mechanism proposed by Tu et al.~\cite{tu2016modeling} that utilizes a coverage vector $c^{t}$, which is a sum of attention vectors at previous time steps as:

\begin{equation}
c^{t} = \sum\limits_{i=0}^{t-1} a^{i}
\end{equation}

At time step $t=1$, when no input has been covered, $c^0$ represents the zero vector. Then, first, we utilize $c^t$ as an extra input to the attention mechanism (discussed in~\ref{s:KeyAtten}) and then modify Equation~\ref{eq:kpvector} as:

\begin{equation}
e^t = {v^T}{tanh({W_h}{h_k} + {W_s}{s_t} + {W_c}{c_{k}^{t}} + b_{att})}
\end{equation}

where $W_c$ is a learnable parameter. Second, to calculate the final loss $L_t$, we follow See et al.~\cite{see2017get} to add negative likelihood loss with $c^{t}$ (as an additional loss) by a hyperparameter $\lambda$ as: 

\begin{equation}
L_{t} = - logP(w_{t} | w_{<t}) + \lambda \sum_{i=0}^{k} min(a_{i}^{t} . c_{i}^{t})
\end{equation}

By incorporating $c^{t}$, the attention module is encouraged to distribute attention weights such that the input words with significant attention through previous decoding steps are assigned lower weights. This encourages the model to explore and focus on different aspects of the input sequence during the generation process.

\subsubsection{Decoding} \label{s:Dec}
\par We use a Beam search algorithm~\cite{tillmann2003word} to create the output summary. While evaluating AuxPGN using test data, we do not provide AuxPGN with any key-phrase information from the key-phrase model. As a result, the right side of Equation~\ref{eq:function} simplifies to the softmax ($e^t$) only. We believe that AuxPGN parameters are optimized during training to maximize the effectiveness of the provided key-phrase attention distribution. This implicit learning enables AuxPGN to detect patterns in the weights of key-related terms, allowing it to utilize the key-related information during summary creation more effectively. Now, we give a set of input tweets and corresponding key-phrases with scores into the AuxPGN encoder model, and then, the decoder model of AuxPGN generates the more human-interpretable summary efficiently. 

\section{Experiments}\label{s:expt}
\par In this Section, we initially discuss the different existing state-of-the-art abstractive summarization approaches that we use as baselines, followed by the experimental setting, comparison results, ablation experiments, comparison with fine-tuned transformer models, and finally, human evaluation experiments.

\subsection{Baselines} \label{s:base} 
\par In this Subsection, we introduce a number of state-of-the-art abstractive summarization approaches, which we use as baselines. 
We segregate these approaches into disaster-specific and non-disaster-specific approaches. We select a few prominent tweet summarization approaches from each type, which we discuss next. 

\begin{enumerate}
    \item Disaster-specific Approaches: Disaster-specific approaches are ones that are specifically developed to summarize disaster-related tweets. From this category, we use the following approaches as baselines to compare with our proposed ATSumm. 
    
        \begin{enumerate}
            \item \textit{CowABS:} Rudra et al.~\cite{rudra2019summarizing} propose an extractive-abstractive framework in which they extract the most important tweets by maximizing the disaster information content of the tweets during the extractive phase. They then form new sentences by traversing the nodes of a bigram-based word graph created from these important tweets. Finally, they incorporate the new sentences into the summary to ensure maximum disaster information coverage and diversity. 
    
            
            \item \textit{CLiQS\_CM:} Vitiugin et al.~\cite{vitiugin2022cross} propose an approach for disaster event summarization, where in extractive phase, they utilize a K-mean clustering algorithm to select most relevant tweets and then in abstractive phase, they utilize a transformer-based language model (i.e., T5~\cite{raffel2020exploring} model) to create the final summary.

            \item \textit{PTESumm:} Rudrapal et al.~\cite{rudrapal2019new} propose an abstractive disaster summarization approach that initially identifies the most relevant and informative tweets using a ranking-based approach and then utilizes a PGN model with coverage technique for abstractive summary generation.  
        \end{enumerate}

    \item Non-disaster-specific Approaches: Non-disaster-specific approaches are ones that are not specifically developed to summarize disaster events but are developed for other summarization domains. From this category, we use the following approaches as baselines to compare with our proposed ATSumm.
    \begin{enumerate}
    
        \item \textit{PGN:} See et al.~\cite{see2017get} propose a pointer generator model that solves the out-of-vocabulary problem by allowing the summarizer to both copy words from the source text and to generate new text.
    
        \item \textit{PSumm:} Zogan et al.~\cite{zogan2021depressionnet} propose an abstractive summarization approach that uses a k-mean algorithm to identify the different clusters and then selects cluster centre as a representative tweet from each cluster. They then construct a concise summary using a pre-trained BART model fine-tuned on the CNN/Daily Mail dataset.   
    
        \item \textit{T5:} Raffel et al.~\cite{raffel2020exploring} propose a pre-trained Seq2Seq transformer transformer-based architecture, T5, for abstractive summarization. For training, they use the Colossal Clean Crawled Corpus. 
        
        \item \textit{BART:} Lewis et al.~\cite{lewis2020bart} propose a pre-trained Seq2Seq encoder-decoder transformer-based architecture, BART, for abstractive summarization. BART's concept is to use the denoising objective during the pre-training stage. 
        
        \item \textit{PEGASUS:} Zhang et al.~\cite{zhang2020pegasus} utilizes a novel summarization-specific pre-training objective known as Gap Sentence Generation (GSG), which involves masking out important sentences from a document and then creating those sentences from the remaining sentences. This training objective helps the model learn how to identify and summarize the most important information in a document. 
        
        \item \textit{Longformer:} Beltagy et al.~\cite{beltagy2020longformer} propose a modified transformer-based Seq2Seq architecture that utilizes a local windowed attention mechanism to address the long input sequence issue. The advantage of this mechanism is it only attends to a small window of tokens at a time, significantly reducing the time it takes to compute the attention weights.  
        
        \item \textit{ProphetNet:} Qi et al.~\cite{qi2020prophetnet} propose PropheNet to address the problem of existing Seq2Seq models failing to capture long-term dependencies between tokens while creating the summary. PropheNet utilizes a novel self-supervised objective known as future n-gram prediction, which encourages the model to learn long-range dependencies between tokens and generate more fluent and grammatically correct summaries.
    \end{enumerate}
\end{enumerate}

\begin{table*}
    \centering 
    \caption{Table shows F1-score (F1) of ROUGE-1, 2, and L, as R-1, R-2, and R-L, respectively, for ATSumm and baselines on $D_1-D_{6}$ disaster datasets. [Note: we mark the best-performing baseline in blue colour.]}
    \label{table:Result1}
    \resizebox{\textwidth}{!}{\begin{tabular}{|c|l|c|c|c|c|l|c|c|c|c|l|c|c|c|c|} \hline
    
        \textbf{Dataset} & \textbf{Approach} & \textbf{R-1} & \textbf{R-2} & \textbf{R-L}  & \textbf{Dataset} & \textbf{Approach} & \textbf{R-1} & \textbf{R-2} & \textbf{R-L} & \textbf{Dataset} & \textbf{Approach} & \textbf{R-1} & \textbf{R-2} & \textbf{R-L} \\ \cline{3-5}\cline{8-10} \cline{12-15}
        
        &  & \textbf{F1} & \textbf{F1} & \textbf{F1} & & & \textbf{F1} & \textbf{F1} & \textbf{F1} & & & \textbf{F1} & \textbf{F1} & \textbf{F1}\\ \hline

                 & $ATSumm$   & \bf0.58 & \bf0.30 & \bf0.41 &           & $ATSumm$     & \bf0.54 & \bf0.25 & \bf0.33 &  & $ATSumm$   & \bf0.42 & \bf0.23 & \bf0.29 \\ \cline{2-5}\cline{7-10} \cline{12-15}
                 & $ExAbsSumm$  & 0.38 & 0.16 & 0.34 &                  & $ExAbsSumm$  & 0.35 & 0.13 & 0.24 &  & $ExAbsSumm$  & 0.31 & 0.06 & 0.24 \\ \cline{2-5}\cline{7-10}\cline{12-15}
                 & $PGN$        & 0.34 & 0.15 & 0.31 &                  & $PGN$        & 0.26 & 0.08 & 0.12 & & $PGN$        & 0.24 & 0.04 & 0.18 \\ \cline{2-5}\cline{7-10}\cline{12-15}
                 & $PTESumm$    & 0.42 & 0.17 & 0.32 &                  & $PTESumm$    & 0.35 & 0.12 & 0.23 & & $PTESumm$    & 0.28 & 0.07 & 0.19 \\ \cline{2-5}\cline{7-10}\cline{12-15}
                 & $PSumm$      & 0.52 & 0.20 & 0.31 &                  & $PSumm$      & 0.46 & 0.22 & 0.29 & & $PSumm$      & 0.37 & 0.19 & 0.26 \\ \cline{2-5}\cline{7-10}\cline{12-15}
                 & $CLiQS\_CM$  & 0.53 & 0.23 & 0.32 &                  & $CLiQS\_CM$  & 0.46 & 0.19 & 0.28 & & $CLiQS\_CM$  & 0.35 & 0.18 & 0.25 \\ \cline{2-5}\cline{7-10}\cline{12-15}
        ${D_1}$  & $T5$ & \blue{0.54} & \blue{0.24} & \blue{0.36}& ${D_5}$ & $T5$      & \blue{0.48} & \blue{0.23} &\blue{ 0.31} & ${D_9}$ & $T5$         & \blue{0.37} & \blue{0.20} & \blue{0.26} \\ \cline{2-5}\cline{7-10}\cline{12-15}
                 & $BART$       & 0.53 & 0.21 & 0.32 &                  & $BART$       & 0.47 & 0.22 & 0.26 & & $BART$       & 0.36 & 0.18 & 0.26 \\ \cline{2-5}\cline{7-10}\cline{12-15}
                 & $PEGASUS$    & 0.39 & 0.12 & 0.29 &                  & $PEGASUS$    & 0.45 & 0.21 & 0.30 &  & $PEGASUS$    & 0.31 & 0.06 & 0.20 \\ \cline{2-5}\cline{7-10}\cline{12-15}
                 & $Longformer$ & 0.50 & 0.22 & 0.36 &                  & $Longformer$ & 0.39 & 0.18 & 0.25 &  & $Longformer$ & 0.34 & 0.09 & 0.25 \\ \cline{2-5}\cline{7-10}\cline{12-15}
                 & $ProphetNet$ & 0.51 & 0.23 & 0.34 &                  & $ProphetNet$ & 0.42 & 0.13 & 0.22 &  & $ProphetNet$ & 0.32 & 0.04 & 0.18 \\ \hline

                 & $ATSumm$   & \bf0.44 & \bf0.20 & \bf0.32 &           & $ATSumm$   & \bf0.50 & \bf0.20 & \bf0.29 &  & $ATSumm$     & \bf0.46 & \bf0.26 & \bf0.36 \\ \cline{2-5}\cline{7-10}\cline{12-15}
                 & $ExAbsSumm$  & 0.17 & 0.06 & 0.16 &                  & $ExAbsSumm$  & 0.33 & 0.07 & 0.19 &   & $ExAbsSumm$  & 0.40 & 0.09 & 0.23 \\ \cline{2-5}\cline{7-10}\cline{12-15}
                 & $PGN$        & 0.25 & 0.05 & 0.15 &                  & $PGN$        & 0.28 & 0.09 & 0.21 &   & $PGN$        & 0.36 & 0.07 & 0.21 \\ \cline{2-5}\cline{7-10}\cline{12-15}
                 & $PTESumm$    & 0.32 & 0.09 & 0.18 &                  & $PTESumm$    & 0.38 & 0.10 & 0.22 &   & $PTESumm$    & 0.40 & 0.15 & 0.31 \\ \cline{2-5}\cline{7-10}\cline{12-15}
                 & $PSumm$      & 0.34 & 0.11 & 0.22 &                  & $PSumm$      & 0.43 & 0.11 & 0.27 &   & $PSumm$      & 0.41 & 0.15 & 0.32 \\ \cline{2-5}\cline{7-10}\cline{12-15}
                 & $CLiQS\_CM$  & 0.32 & 0.11 & 0.25 &                  & $CLiQS\_CM$  & 0.41 & 0.11 & 0.25 &   & $CLiQS\_CM$  & 0.42 & 0.18 & 0.30 \\ \cline{2-5}\cline{7-10}\cline{12-15}
         ${D_2}$ & $T5$         & 0.40 & 0.13 & 0.25 &     ${D_6}$      & $T5$         & \blue{0.44} & \blue{0.12} & \blue{0.27} &  ${D_{10}}$ & $T5$         & \blue{0.43} & \blue{0.22} & \blue{0.33} \\ \cline{2-5}\cline{7-10}\cline{12-15}
                 & $BART$       & 0.31 & 0.04 & 0.19 &                  & $BART$       & 0.42 & 0.13 & 0.25 &  & $BART$       & 0.40 & 0.16 & 0.31 \\ \cline{2-5}\cline{7-10}\cline{12-15}
                 & $PEGASUS$    & 0.32 & 0.06 & 0.19 &                  & $PEGASUS$    & 0.43 & 0.12 & 0.24 &   & $PEGASUS$    & 0.26 & 0.07 & 0.18 \\ \cline{2-5}\cline{7-10}\cline{12-15}
                 & $Longformer$ & \blue{0.40} & \blue{0.14} & \blue{0.26} &  & $Longformer$ & 0.39 & 0.10 & 0.21 &  & $Longformer$ & 0.42 & 0.21 & 0.30                  \\ \cline{2-5}\cline{7-10}\cline{12-15}
                 & $ProphetNet$ & 0.36 & 0.05 & 0.18 &                  & $ProphetNet$ & 0.39 & 0.11 & 0.18 &   & $ProphetNet$ & 0.41 & 0.21 & 0.31 \\ \hline

                 & $ATSumm$   & \bf0.48 & \bf0.24 & \bf0.30 &           & $ATSumm$   & \bf0.48 & \bf0.21 & \bf0.29 & & $ATSumm$     & \bf0.55 & \bf0.31 & \bf0.33 \\ \cline{2-5}\cline{7-10}\cline{12-15}
                 & $ExAbsSumm$  & 0.24 & 0.06 & 0.22 &                  & $ExAbsSumm$  & 0.35 & 0.07 & 0.23 & & $ExAbsSumm$  & 0.44 & 0.12 & 0.26 \\ \cline{2-5}\cline{7-10}\cline{12-15}
                 & $PGN$        & 0.23 & 0.06 & 0.21 &                  & $PGN$        & 0.31 & 0.09 & 0.24 & & $PGN$        & 0.39 & 0.07 & 0.23 \\ \cline{2-5}\cline{7-10}\cline{12-15}
                 & $PTESumm$    & 0.35 & 0.08 & 0.22 &                  & $PTESumm$    & 0.40 & 0.11 & 0.25 & & $PTESumm$    & 0.42 & 0.11 & 0.24 \\ \cline{2-5}\cline{7-10}\cline{12-15}
                 & $PSumm$      & 0.41 & 0.10 & 0.26 &                  & $PSumm$      & 0.46 & 0.15 & 0.27 & & $PSumm$      & 0.50 & 0.14 & 0.26 \\ \cline{2-5}\cline{7-10}\cline{12-15}
                 & $CLiQS\_CM$  & 0.41 & 0.15 & 0.26 &                  & $CLiQS\_CM$  & 0.43 & 0.12 & 0.25 & & $CLiQS\_CM$  & 0.51 & 0.20 & 0.26 \\ \cline{2-5}\cline{7-10}\cline{12-15}
         ${D_3}$ & $T5$         & 0.40 & 0.14 & 0.25 &         ${D_7}$  & $T5$         & 0.40 & 0.09 & 0.23 & ${D_{11}}$ & $T5$         & 0.52 & 0.22 & 0.25 \\ \cline{2-5}\cline{7-10}\cline{12-15}
                 & $BART$       & 0.40 & 0.09 & 0.25 &                  & $BART$       & \blue{0.46} & \blue{0.16} & \blue{0.28} &  & $BART$       & \blue{0.53} & \blue{0.23} & \blue{0.28} \\ \cline{2-5}\cline{7-10}\cline{12-15}
                 & $PEGASUS$    & 0.39 & 0.07 & 0.23 &                  & $PEGASUS$    & 0.44 & 0.14 & 0.25 & & $PEGASUS$    & 0.43 & 0.11 & 0.26 \\ \cline{2-5}\cline{7-10}\cline{12-15}
                 & $Longformer$ & 0.39 & 0.15 & 0.27 &                  & $Longformer$ & 0.46 & 0.16 & 0.26 & & $Longformer$ & 0.41 & 0.15 & 0.26 \\ \cline{2-5}\cline{7-10}\cline{12-15}
                 & $ProphetNet$ & \blue{0.42} & \blue{0.15} & \blue{0.28} &  & $ProphetNet$ & 0.29 & 0.11 & 0.24 & & $ProphetNet$ & 0.37 & 0.06 & 0.18 \\ \hline

                 & $ATSumm$ & \bf0.52 & \bf0.28 & \bf0.34 & & $ATSumm$   & \bf0.51 & \bf0.26 & \bf0.30 &  & $ATSumm$   & \bf0.49 & \bf0.24 & \bf0.34 \\ \cline{2-5}\cline{7-10}\cline{12-15}
                 & $ExAbsSumm$  & 0.28 & 0.10 & 0.23 &      & $ExAbsSumm$  & 0.40 & 0.10 & 0.25 & & $ExAbsSumm$  & 0.34 & 0.10 & 0.23 \\ \cline{2-5}\cline{7-10}\cline{12-15}
                 & $PGN$        & 0.31 & 0.09 & 0.26 &      & $PGN$        & 0.33 & 0.06 & 0.22 & & $PGN$        & 0.37 & 0.14 & 0.27 \\ \cline{2-5}\cline{7-10}\cline{12-15}
                 & $PTESumm$    & 0.38 & 0.15 & 0.21 &      & $PTESumm$    & 0.38 & 0.11 & 0.21 & & $PTESumm$    & 0.39 & 0.10 & 0.22     \\ \cline{2-5}\cline{7-10}\cline{12-15}
                 & $PSumm$      & 0.44 & 0.25 & 0.32 &      & $PSumm$      & 0.45 & 0.17 & 0.25 & & $PSumm$      & \blue{0.44} & \blue{0.17} & \blue{0.29} \\\cline{2-5}\cline{7-10}\cline{12-15}
                 & $CLiQS\_CM$  & 0.41 & 0.24 & 0.30 &      & $CLiQS\_CM$  & 0.41 & 0.15 & 0.24 & & $CLiQS\_CM$ & 0.39 & 0.15 & 0.27 \\\cline{2-5}\cline{7-10}\cline{12-15}
        ${D_4}$  & $T5$         & 0.43 & 0.24 & 0.31 & ${D_8}$ & $T5$  & 0.39 & 0.12 & 0.23 &  ${D_{12}}$ & $T5$         & 0.40 & 0.17 & 0.29\\\cline{2-5}\cline{7-10}\cline{12-15}
                 & $BART$       & 0.44 & 0.25 & 0.31 &      & $BART$   & \blue{0.47} &\blue{ 0.17} & \blue{0.28} & & $BART$       & 0.40 & 0.16 & 0.27 \\\cline{2-5}\cline{7-10}\cline{12-15}
                 & $PEGASUS$    & \blue{0.45} & \blue{0.26} & \blue{0.32} &  & $PEGASUS$    & 0.37 & 0.08 & 0.23 & & $PEGASUS$    & 0.36 & 0.08 & 0.21 \\\cline{2-5}\cline{7-10}\cline{12-15}
                 & $Longformer$ & 0.39 & 0.20 & 0.24 &      & $Longformer$ & 0.45 & 0.16 & 0.25 & & $Longformer$ & 0.44 & 0.16 & 0.26 \\\cline{2-5}\cline{7-10}\cline{12-15}
                 & $ProphetNet$ & 0.41 & 0.18 & 0.26 &      & $ProphetNet$ & 0.39 & 0.10 & 0.24 & & $ProphetNet$ & 0.40 & 0.09 & 0.22 \\ \hline

                & $ATSumm$     & \bf0.52 & \bf0.25 & \bf0.35 & \multicolumn{5}{c}{}\\ \cline{2-5}
                 & $ExAbsSumm$  & 0.43 & 0.15 & 0.30 & \multicolumn{5}{c}{}\\ \cline{2-5}
                 & $PGN$        & 0.37 & 0.07 & 0.20 & \multicolumn{5}{c}{}\\ \cline{2-5}
                 & $PTESumm$    & 0.40 & 0.09 & 0.23 & \multicolumn{5}{c}{}\\ \cline{2-5}
                 & $PSumm$      & 0.44 & 0.17 & 0.30 & \multicolumn{5}{c}{}\\ \cline{2-5}
                 & $CLiQS\_CM$  & 0.45 & 0.20 & 0.29 & \multicolumn{5}{c}{}\\ \cline{2-5}
      ${D_{13}}$ & $T5$         & \blue{0.47} & \blue{0.22} & \blue{0.31} & \multicolumn{5}{c}{}\\ \cline{2-5}
                 & $BART$       & 0.42 & 0.14 & 0.25 & \multicolumn{5}{c}{}\\ \cline{2-5}
                 & $PEGASUS$    & 0.43 & 0.14 & 0.29 & \multicolumn{5}{c}{}\\ \cline{2-5}
                 & $Longformer$ & 0.41 & 0.18 & 0.25 & \multicolumn{5}{c}{}\\ \cline{2-5}
                 & $ProphetNet$ & 0.39 & 0.08 & 0.22 & \multicolumn{5}{c}{}\\ \cline{1-5}
    \end{tabular} }
\end{table*}

\subsection{Experimental Setting} \label{s:expset}
\par For our hyperparameters, we use the hidden state dimensions of the LSTMs to $256$, the word-embedding dimensions at $128$, and the min-batch size to $16$. For AuxPGN model, we have set the input tweet length to $400$ words while the output summary length to $200$ words. 
For training, we use Adagard~\cite{duchi2011adaptive} with a learning rate of $0.15$ and an initial accumulator value of $0.1$. Furthermore, when decoding the summaries for the test dataset, we use the beam size to $5$ and the minimum generating summary length to $35$ tokens. We also restricted the vocabulary of our model to $50,000$ tokens. We run these experiments on a Google Collab GPU environment using Keras and TensorFlow with a Nvidia K80/T4 GPU with $25$ GB RAM. We set these hyperparameters through the various experiments discussed in Subsection~\ref{s:hyp}.

\par We follow the same procedure as Poddar et al.~\cite{poddar2022caves} to construct an abstractive summary using pre-trained transformer models. We initially split the large-sized input tweets into equal-sized chunks/samples so that the length of each sample did not exceed the maximum permissible length. Then, we obtain the summary for each sample and combine these sampled summaries to form the overall summary. Further, we create a final summary of $200$ words in length by selecting top-ranked sentences (from the combined summary) based on their TF-IDF scores.

\subsection{Comparison with Existing Research Works}  \label{s:res}
\par To evaluate the summary generated by ATSumm and different baselines with the gold standard summaries, we use the ROUGE-N score~\cite{lin2004rouge}. ROUGE-N calculates a score based on the number of words that overlap between the model-generated summary and the gold standard summary. We use the F1-score for three variants of the ROUGE-N scores, where N takes 1, 2, and L values, respectively. The higher ROUGE-N score indicates the higher the quality of the resulting summary. Our observations, as shown in Table~\ref{table:Result1} indicate that the ATSumm outperforms over all the baselines by $3.63-61.36$\%, $7.14-80$\%, and $3.45-63.63$\% in terms of ROUGE-1, 2, and L F1-score, respectively and it above over all the baselines. We found that the highest and lowest improvements are inconsistent with the baselines. For example, the improvement is the highest over PGN for $D_2$, $D_3$, $D_5$, $D_6$, $D_8$, $D_9$, $D_{11}$, and $D_{13}$, CowABS for $D_{4}$, and PEGASUS for $D_{10}$, and $D_{12}$ and lowest over T5 for $D_1$, $D_5$, $D_6$, $D_9$, $D_{10}$, and $D_{13}$ BART for $D_7$, $D_8$, and $D_{11}$, Longformer for $D_2$, ProphetNet for $D_3$, PSumm for $D_{12}$, and PEGASUS for $D_4$. The improvement of ATSumm is the best for $D_1$ with $0.58-0.30$ and worst for $D_9$ with $0.42-0.23$ in terms of ROUGE-N F1-score. We also observe that the improvement compared to the best-performing baseline (marked in blue colour in the tables) in each scenario remains in the range of $0.02-0.07$ in terms of ROUGE-1 F1-score. However, the best-performing baseline is not constant across the datasets. Therefore, there is always a significant margin of good results for ATSumm with the best-performing baseline.

\subsection{Ablation Experiments}\label{s:ablexp}
\par To understand the effectiveness of each phase of ATSumm, we compare the performance of ATSumm with its different variants and study the role of auxiliary information in the proposed AuxPGN model. 

\subsubsection{Phase-I} In this experiment, we compare Phase-I of ATSumm with its different variants, which are as follows:    
\begin{itemize}
    \item In \textit{ATSumm-1A}\footnote{The first number in the name of each of the variants of ATSumm, i.e., \textit{ATSumm-1A} represents the phase, and the second number denotes the sequence of the ablation experiment in that phase.}, we extract the top-ranked tweets based on an ontology-based tweet summarization approach proposed by Garg et al.~\cite{garg2023ontodsumm}. 

    \item In \textit{ATSumm-1B}, we extract the top-ranked tweets based on an entropy-based tweet summarization approach proposed by Garg et al.~\cite{Garg2022Entropy}.

    \item In \textit{ATSumm-1C}, we extract the top-ranked tweets based on the importance of their content words (i.e., nouns, verbs, numerals, and adjectives) using TF-IDF, proposed by Rudra et al.~\cite{rudra2015extracting}. 
\end{itemize}

\begin{table*}
    \centering 
    \caption{Table shows F1-score (F1) of ROUGE-1, 2, and L, as R-1, R-2, and R-L, respectively, for Phase-I and Phase-II variants of ATSumm, i.e.,  
    \textit{ATSumm-1A}, \textit{ATSumm-1B}, \textit{ATSumm-1C}, \textit{ATSumm-2A}, \textit{ATSumm-2B}, \textit{ATSumm-2C}, \textit{ATSumm-2D}, \textit{ATSumm-2E}, \textit{ATSumm-2F}, and \textit{ATSumm-2G} with ATSumm on six datasets: $D_1$, $D_3$, $D_7$, $D_8$, $D_{10}$, and $D_{12}$.}
    \label{table:abal}
    \resizebox{0.95\textwidth}{!}{\begin{tabular}{|c|l|c|c|c|c|l|c|c|c|c|l|c|c|c|} \hline
        \textbf{Dataset} & \textbf{Approach} & \textbf{R-1} & \textbf{R-2} & \textbf{R-L} & \textbf{Dataset} & \textbf{Approach} & \textbf{R-1} & \textbf{R-2} & \textbf{R-L} & \textbf{Dataset} & \textbf{Approach} & \textbf{R-1} & \textbf{R-2} & \textbf{R-L} \\ \cline{3-5} \cline{7-10} \cline{12-15}
        
        & \textbf{} & \textbf{F1} & \textbf{F1} & \textbf{F1} & & \textbf{} & \textbf{F1} & \textbf{F1} & \textbf{F1} & & \textbf{} & \textbf{F1} & \textbf{F1} & \textbf{F1}\\ \hline

                & $ATSumm$    & \bf0.58 & \bf0.30 & \bf0.41 &      & $ATSumm$    & \bf0.48 & \bf0.21 & \bf0.29 &        & $ATSumm$    & \bf0.46 & \bf0.26 & \bf0.36 \\ \cline{2-5} \cline{7-10} \cline{12-15}
                & $ATSumm-1A$ & 0.50 & 0.27 & 0.33 &               & $ATSumm-1A$ & 0.43 & 0.15 & 0.27 &                 & $ATSumm-1A$ & 0.43 & 0.24 & 0.35 \\ \cline{2-5} \cline{7-10}\cline{12-15}
                & $ATSumm-1B$ & 0.42 & 0.28 & 0.35 &               & $ATSumm-1B$ & 0.35 & 0.07 & 0.21 &                 & $ATSumm-1B$ & 0.42 & 0.21 & 0.31 \\ \cline{2-5} \cline{7-10}\cline{12-15}
                & $ATSumm-1C$ & 0.38 & 0.11 & 0.26 &               & $ATSumm-1C$ & 0.32 & 0.11 & 0.22 &                 & $ATSumm-1C$ & 0.37 & 0.14 & 0.29 \\ \cline{2-5} \cline{7-10} \cline{12-15}
                & $ATSumm-2A$ & 0.33 & 0.18 & 0.27 &               & $ATSumm-2A$ & 0.43 & 0.13 & 0.22 &                 & $ATSumm-2A$ & 0.43 & 0.21 & 0.32 \\ \cline{2-5} \cline{7-10} \cline{12-15}
    {${D_1}$}& $ATSumm-2B$ & 0.47 & 0.18 & 0.30 &  {${D_7}$} & $ATSumm-2B$ & 0.44 & 0.16 & 0.23 & {${D_{10}}$} & $ATSumm-2B$ & 0.42 & 0.23 & 0.31 \\ \cline{2-5} \cline{7-10} \cline{12-15}
                & $ATSumm-2C$ & 0.34 & 0.07 & 0.23 &               & $ATSumm-2C$ & 0.26 & 0.05 & 0.19 &                 & $ATSumm-2C$ & 0.32 & 0.12 & 0.22 \\ \cline{2-5} \cline{7-10} \cline{12-15}
                & $ATSumm-2D$ & 0.44 & 0.18 & 0.35 &               & $ATSumm-2D$ & 0.45 & 0.15 & 0.26 &                 & $ATSumm-2D$ & 0.36 & 0.19 & 0.27 \\ \cline{2-5} \cline{7-10} \cline{12-15}
                & $ATSumm-2E$ & 0.28 & 0.09 & 0.23 &               & $ATSumm-2E$ & 0.36 & 0.09 & 0.25 &                 & $ATSumm-2E$ & 0.40 & 0.18 & 0.25 \\ \cline{2-5} \cline{7-10} \cline{12-15}
                & $ATSumm-2F$ & 0.31 & 0.13 & 0.27 &               & $ATSumm-2F$ & 0.34 & 0.13 & 0.24 &                 & $ATSumm-2F$ & 0.39 & 0.20 & 0.21 \\ \cline{2-5} \cline{7-10} \cline{12-15}
                & $ATSumm-2G$ & 0.56 & 0.28 & 0.39 &               & $ATSumm-2G$ & 0.46 & 0.19 & 0.27 &                 & $ATSumm-2G$ & 0.44 & 0.23 & 0.34 \\ \hline

                & $ATSumm$    & \bf0.48 & \bf0.24 & \bf0.30 &      & $ATSumm$    & \bf0.51 & \bf0.26 & \bf0.30 &          & $ATSumm$    & \bf0.49 & \bf0.24 & \bf0.34 \\ \cline{2-5} \cline{7-10} \cline{12-15}  
                & $ATSumm-1A$ & 0.42 & 0.14 & 0.27 &               & $ATSumm-1A$ & 0.49 & 0.22 & 0.24 & & $ATSumm-1A$ & 0.44 & 0.20 & 0.28  \\ \cline{2-5} \cline{7-10} \cline{12-15}       
                & $ATSumm-1B$ & 0.37 & 0.19 & 0.29 &               & $ATSumm-1B$ & 0.42 & 0.16 & 0.26 & & $ATSumm-1B$ & 0.38 & 0.12 & 0.28 \\ \cline{2-5} \cline{7-10} \cline{12-15}      
                & $ATSumm-1C$ & 0.35 & 0.07 & 0.23 &               & $ATSumm-1C$ & 0.40 & 0.16 & 0.27 & & $ATSumm-1C$ & 0.37 & 0.14 & 0.28 \\ \cline{2-5} \cline{7-10} \cline{12-15}     
                & $ATSumm-2A$ & 0.31 & 0.17 & 0.27 &               & $ATSumm-2A$ & 0.41 & 0.17 & 0.24 & & $ATSumm-2A$ & 0.41 & 0.13 & 0.25 \\ \cline{2-5} \cline{7-10} \cline{12-15}     
    {${D_3}$}& $ATSumm-2B$ & 0.40 & 0.15 & 0.26 & {{${D_8}$}}& $ATSumm-2B$ & 0.47 & 0.18 & 0.28 & {${D_{12}}$}& $ATSumm-2B$ & 0.43 & 0.15 & 0.27 \\ \cline{2-5} \cline{7-10} \cline{12-15}  
                & $ATSumm-2C$ & 0.30 & 0.04 & 0.22 &               & $ATSumm-2C$ & 0.19 & 0.05 & 0.18 & & $ATSumm-2C$ & 0.21 & 0.05 & 0.16 \\ \cline{2-5} \cline{7-10} \cline{12-15}     
                & $ATSumm-2D$ & 0.35 & 0.16 & 0.27 &               & $ATSumm-2D$ & 0.45 & 0.14 & 0.25 & & $ATSumm-2D$ & 0.44 & 0.17 & 0.29 \\ \cline{2-5} \cline{7-10} \cline{12-15}       
                & $ATSumm-2E$ & 0.26 & 0.03 & 0.18 &               & $ATSumm-2E$ & 0.38 & 0.13 & 0.27 & & $ATSumm-2E$ & 0.25 & 0.10 & 0.23 \\ \cline{2-5} \cline{7-10} \cline{12-15}   
                & $ATSumm-2F$ & 0.28 & 0.14 & 0.17 &               & $ATSumm-2F$ & 0.36 & 0.13 & 0.25 & & $ATSumm-2F$ & 0.34 & 0.15 & 0.22 \\ \cline{2-5} \cline{7-10} \cline{12-15}       
                & $ATSumm-2G$ & 0.45 & 0.21 & 0.27 &               & $ATSumm-2G$ & 0.48 & 0.24 & 0.27 & & $ATSumm-2G$ & 0.47 & 0.22 & 0.31 \\ \hline       
                                          
    \end{tabular}} 
\end{table*}

\subsubsection{Phase-II} To understand the role and effectiveness of the AuxPGN model in Phase-II of ATSumm, we compare the performance of ATSumm with its different variants, which are as follows:    
\begin{itemize}
    \item In \textit{ATSumm-2A}, we use the T5-small model proposed by Raffel et al.~\cite{raffel2020exploring} to create an abstractive summary of the input tweets. 

    \item In \textit{ATSumm-2B}, we use the BART model proposed by Lewis et al.~\cite{lewis2020bart} to create an abstractive summary of the input tweets.

    \item In \textit{ATSumm-2C}, we use the PEGASUS model proposed by Zhang et al.~\cite{zhang2020pegasus} to create an abstractive summary of the input tweets.

    \item In \textit{ATSumm-2D}, we use the Longformer model proposed by Beltagy et al.~\cite{beltagy2020longformer} to create an abstractive summary of the input tweets.

    \item In \textit{ATSumm-2E}, we use the ProphetNet model proposed by Qi et al.~\cite{qi2020prophetnet} to create an abstractive summary of the input tweets.

    \item In \textit{ATSumm-2F}, we use the PGN model proposed by See et al.~\cite{see2017get} to create an abstractive summary of the input tweets.

    \item In \textit{ATSumm-2G}, we use the proposed AuxPGN model without considering the key-phrase scores to create an abstractive summary of the given input tweets.
\end{itemize}

\subsubsection{Results and Discussions}\label{s:result}
\par Now, we compare and evaluate the performance of the discussed different variants of ATSumm on six different disaster event datasets, such as $D_1$, $D_3$, $D_7$, $D_8$, $D_{10}$, and $D_{12}$ on ROUGE-1, 2, and L F1-score. Our observations in Table~\ref{table:abal} indicate that the ATSumm outperforms all its variants by $3.44-62.74\%$, $6.67-87.50\%$, and $3.33-78.26\%$, ROUGE-1, 2, and L F1-score, respectively, across disasters.  

\begin{enumerate}
    \item \textit{Phase-I: } Our observations indicate that ATSumm outperforms \textit{ATSumm-1C}, which extracts top-ranked tweets based on the importance of their content words using TF-IDF by $3.92-34.48\%$, $6.67-70.83$\%, $3.33-36.58$\% in ROUGE-1, 2, and L F1-score. These results highlight the effectiveness of incorporating key-phrases with the BERT model to extract the tweets in the extractive phase.  
    
    \item \textit{Phase-II: } Our observations indicate that the inclusion of AUxPGN in ATSumm outperforms of $3.44-62.74$\%, $6.67-87.50$\%, and $4.87-78.26$\% in ROUGE-1, 2, and L F1-score, respectively, with its different variants, such as \textit{ATSumm-2A}, \textit{ATSumm-2B}, \textit{ATSumm-2C}, \textit{ATSumm-2D}, \textit{ATSumm-2E}, \textit{ATSumm-2F}, and \textit{ATSumm-2G}. This justifies that incorporating the auxiliary information (key-phrases and their importance scores) in the PGN helps to create a more human-interpretable summary even if we provide less training data to train it. We further observe that the information gain of ATSumm is highest over \textit{ATSumm-2C} as it employs the PEGASUS model to generate the summary, while it is lowest over \textit{ATSumm-2G}, which combines only key-phrases in AuxPGN for summary creation. We also observe that the inclusion of novel AuxPGN in Phase-II provides maximum performance of ATSumm. 
\end{enumerate}

\begin{figure*}[!ht]
\centering
\subfloat[]{\includegraphics[width=0.33\linewidth]{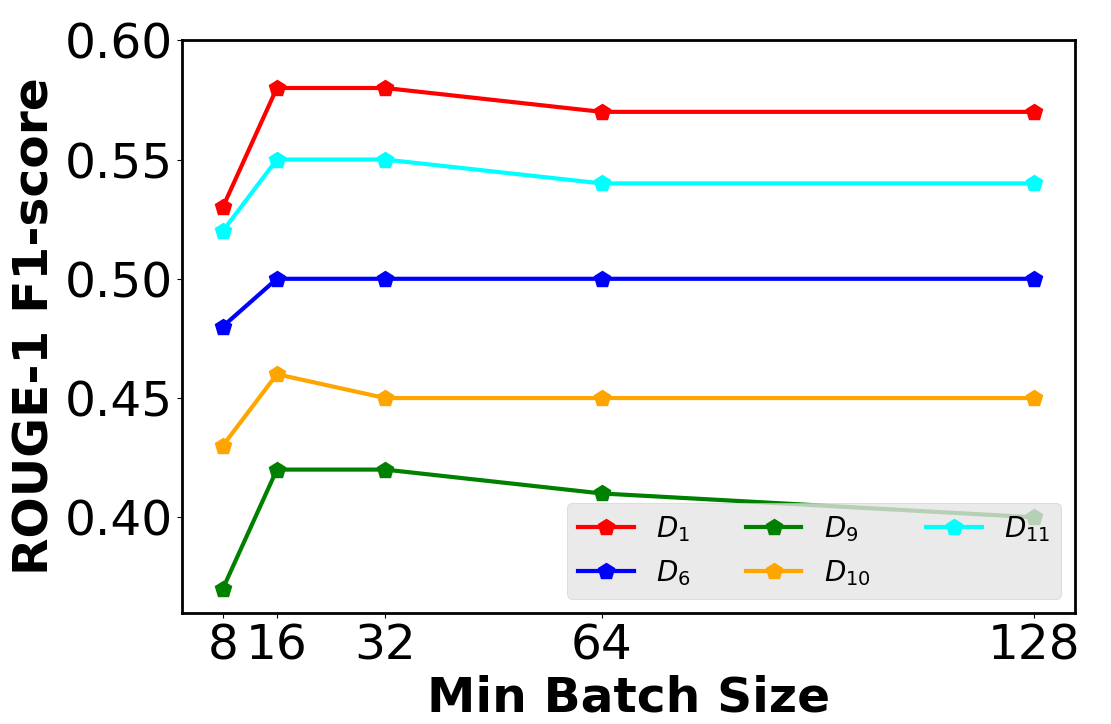} \label{fig:minbatch}}
\subfloat[]{\includegraphics[width=0.33\linewidth]{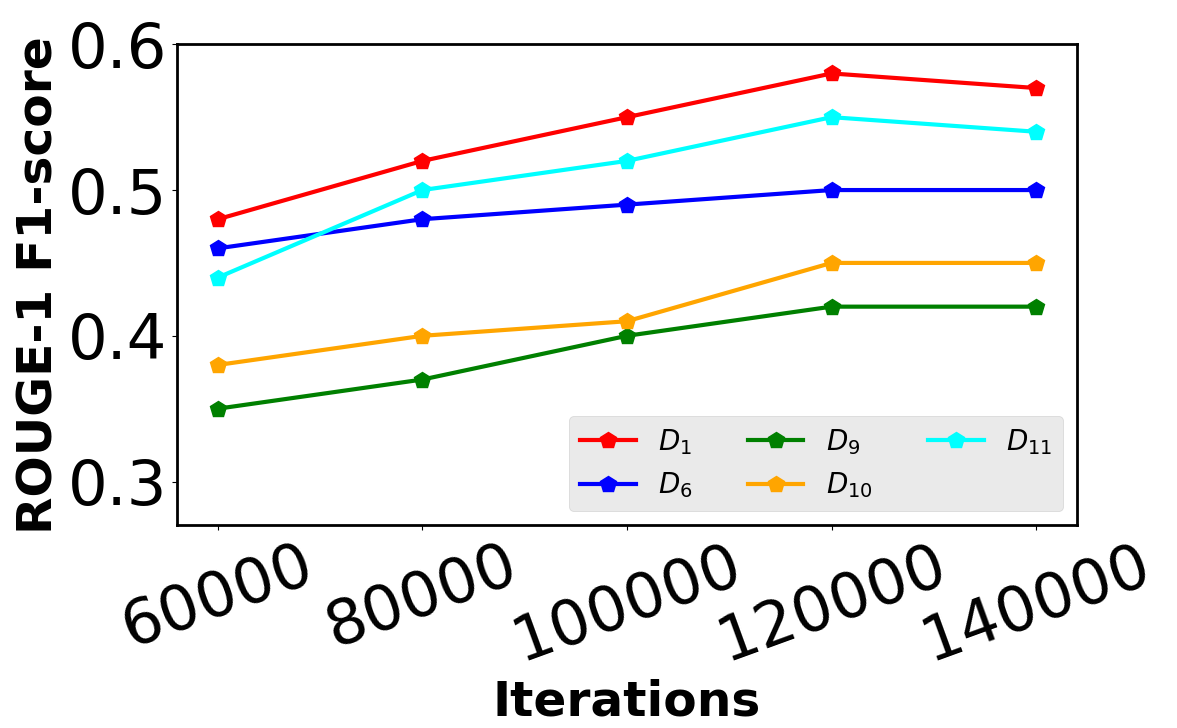} \label{fig:iteration}}
\subfloat[]{\includegraphics[width=0.33\linewidth]{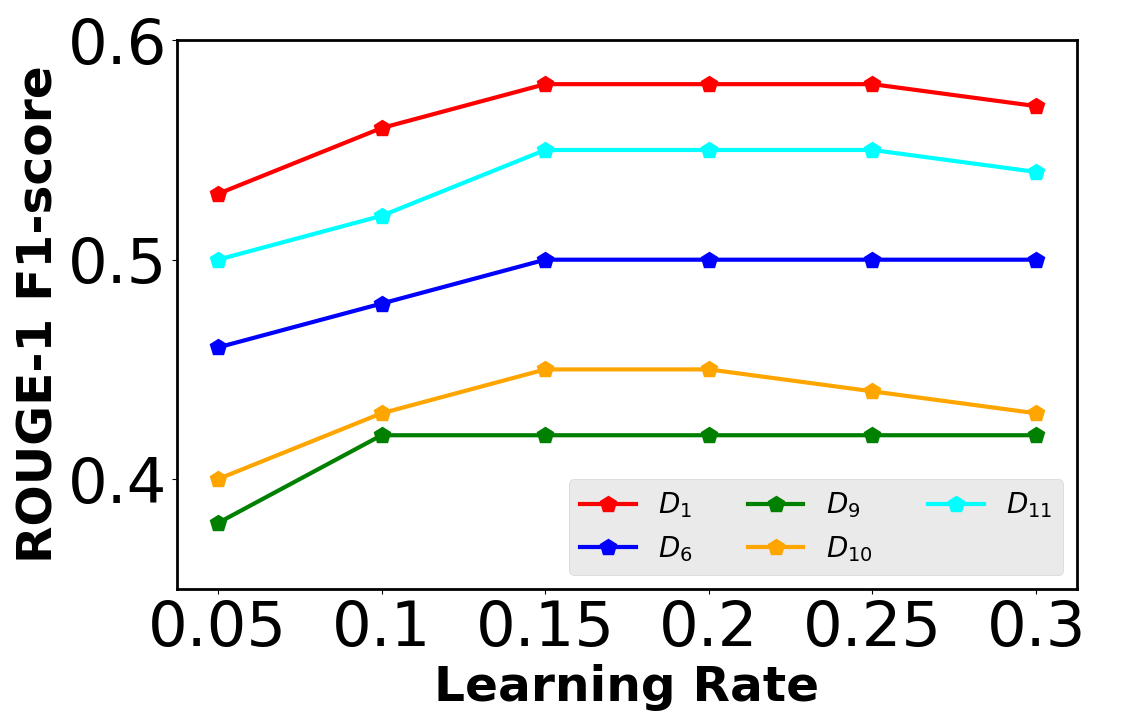} \label{fig:learning}}
\caption{Figure shows the comparison results of ROUGE-1 F1-score of ATSumm with the different hyperparameters, i.e., min batch size, learning rate, and number of training iterations on five datasets, such as $D_1$, $D_6$, $D_9$,  $D_{10}$, and $D_{11}$ in Figure~\ref{fig:minbatch}, Figure~\ref{fig:iteration} and ~\ref{fig:learning}, respectively.}
\label{figure:hyperparameter}
\end{figure*}

\subsection{Hyperparameter Analysis} \label{s:hyp}
\par In this Subsection, we discuss how we set the values of hyperparameters for ATSumm, i.e., min batch size, learning rate, and number of training iterations. For each parameter or hyperparameter, we vary its values and calculate the ROUGE-1 F1-score for each value, and then we select that value for a hyperparameter that gives the best ROUGE-1 F1-score. We randomly select five different datasets for our experiment on different hyperparameters: $D_1$, $D_6$, $D_9$,  $D_{10}$, and $D_{11}$ and the experimental results shown in Figure~\ref{figure:hyperparameter}. Our observations from Figure~\ref{fig:minbatch} indicate that by varying the min batch size from $8$ to $128$, the min batch size of $16$ provides the maximum ROUGE-1 F1-score irrespective of the datasets. We also observe that there is no significant improvement when we increase the min batch size from $32$ to $128$. Therefore, we select $16$ as a min batch size in ATSumm. Similarly, we observe that $1,20,000$ number of iterations provides the maximum ROUGE-1 F1-score for ATSumm across datasets, as shown in Figure~\ref{fig:iteration}. Similarly, to decide the learning rate of Adagrad~\cite{duchi2011adaptive} optimizer, we vary the learning rate from $0.05$ to $0.30$ and measure ROUGE-1 F1-score for ATSumm across five datasets. Our observations from Figure~\ref{fig:learning} indicate that the learning rate of $0.15$ provides the maximum ROUGE-1 F1-score and no significant improvement when we vary the learning rate from $0.20$ to $0.30$. Therefore, in ATSumm, we select $0.15$ as a min learning rate.

\subsection{Scalability Analysis} 
\par
\begin{figure}[ht]
\centering
\includegraphics[width=0.75\linewidth]{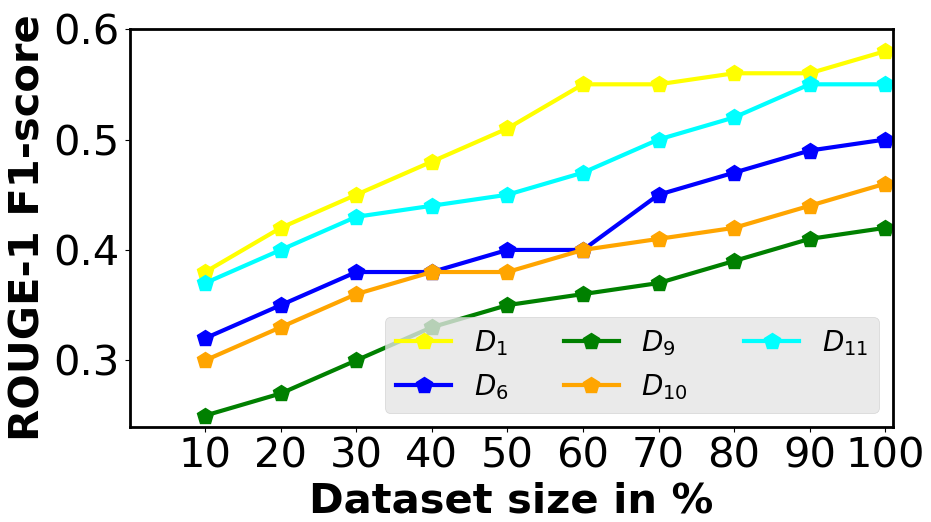} 
\caption{Figure shows the comparison results of ROUGE-1 F1-score of ATSumm with the different sizes of the ARIES training dataset in percentage.}
\label{figure:Scalability}
\end{figure}

 In this Subsection, we investigate the scalability of AuxPGN in ATSumm by comparing its performance when the training dataset size increases gradually from $10$\% to $100$\%. We select five different datasets: $D_1$, $D_6$, $D_9$, $D_{10}$, and $D_{11}$ for our experiment on different training dataset sizes and follow the same procedure for implementation as discussed in Section~\ref{s:prop} and then calculate the ROUGE-1 F1-score. Our results, as shown in Figure~\ref{figure:Scalability}, indicate that AuxPGN is scalable with increasing training dataset size. We also observe that the ATSumm ensure significantly higher performance in terms of ROUGE-1 F1-score for $100$\% training dataset size than other training sizes. Subsequently, ATSumm ensures the lowest ROUGE-1 F1-score with $10$\% training dataset size. Additionally, we observe that after the $90$\% training dataset, the F1-score is almost constant, and there is not much further improvement. Hence, we train the AuxPGN model in ATSumm on $100$\% training samples, which are $7500$ samples of the ARIES dataset.


\subsection{Weight Analysis}\label{s:thre}
\par
\begin{figure}[ht]
\centering
\includegraphics[width=1.0\linewidth]{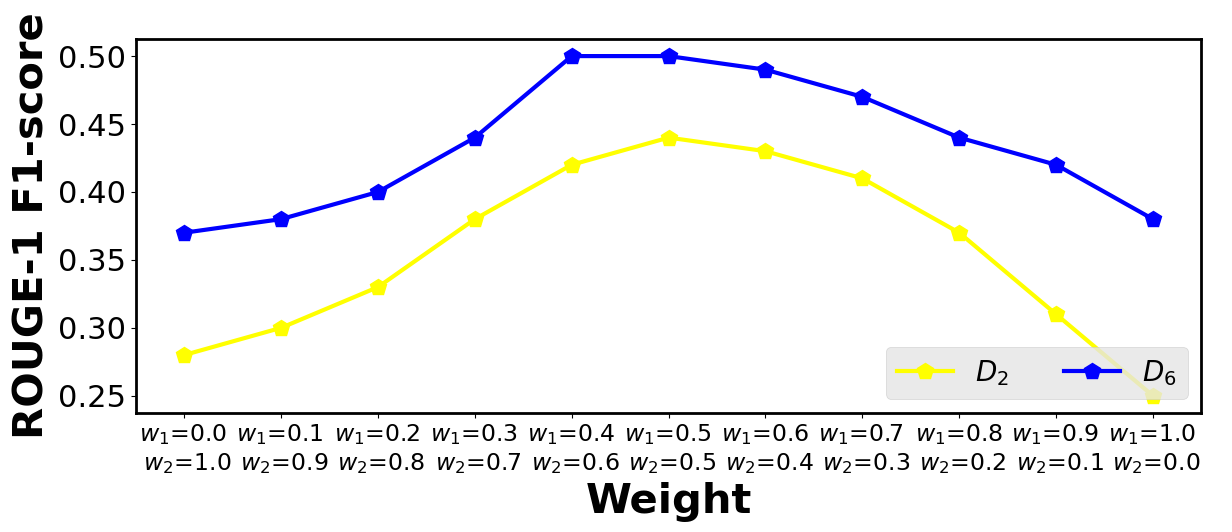} 
\caption{Figure shows the ROUGE-1 F1-score of each value of pair of $w_1$ and $w_2$ for $D_2$ and $D_6$ datasets.}
\label{figure:weight}
\end{figure}

 In this Subsection, we experimentally choose the weight values of $w_1$ and $w_2$ used in Equation~\ref{eq:function}. For our experiment, we randomly select two disaster datasets: $D_2$ and $D_6$ and vary the range of $w_1$ from $0$ to $1$ and $w_2$ from $1$ to $0$ such that $w_1+w_2=1$. For each value of pair of $w_1$ and $w_2$, we train the AuxPGN model and then generate the summary using ATSumm. Further, we evaluate the performance of these generated summaries using ATSumm with the gold standard summaries using ROUGE-1 F1-score. We show the ROUGE-1 F1-score of each value of pair of $w_1$ and $w_2$ for $D_2$ and $D_6$ in Figure~\ref{figure:weight}. Our observations indicate that the ROUGE-1 F1-score is highest when $w_1=w_2=0.5$ compared to other combination values of $w_1$ and $w_2$. Therefore, based on our observations, we select $0.5$ as the weights of $w_1$ and $w_2$ in ATSumm.

\begin{table*}[!ht]
    \centering 
    \caption{Table shows the comparison results of five existing state-of-the-art transformer models, i.e., T5, BART, PEGASUS, Longformer, and ProphetNet which are fine-tuned on our proposed ARIES dataset with ATSumm model for F1-score (F1) of ROUGE-1, 2, and L, as R-1, R-2, and R-L, respectively, on six datasets: $D_1$, $D_3$, $D_7$, $D_8$, $D_{10}$, and $D_{12}$.}
    \label{table:finetune}
    \resizebox{0.8\linewidth}{!}{\begin{tabular}{|c|l|c|c|c|c|l|c|c|c|} \hline
        \textbf{Dataset} & \textbf{Approach} & \textbf{R-1} & \textbf{R-2} & \textbf{R-L} & \textbf{Dataset} & \textbf{Approach} & \textbf{R-1} & \textbf{R-2} & \textbf{R-L} \\ \cline{3-5} \cline{8-10}
        
        & \textbf{} & \textbf{F1} & \textbf{F1} & \textbf{F1} & & \textbf{} & \textbf{F1} & \textbf{F1} & \textbf{F1} \\ \hline

                                & $ATSumm$                   & \bf0.58 & \bf0.30 & \bf0.41 &    & $ATSumm$                   & \bf0.51 & \bf0.26 & \bf0.30 \\ \cline{2-5} \cline{7-10}
                                & $T5-Fine\_tuning$          & 0.55 & 0.26 & 0.38 &             & $T5-Fine\_tuning$           & 0.47 & 0.21 & 0.26 \\ \cline{2-5} \cline{7-10}
    ${D_1}$ & $BART-Fine\_tuning$        & 0.55 & 0.26 & 0.37 &  ${D_8}$  & $BART-Fine\_tuning$ & 0.49 & 0.22 & 0.28 \\ \cline{2-5} \cline{7-10}
                                & $PEGASUS-Fine\_tuning$     & 0.48 & 0.24 & 0.33 &             & $PEGASUS-Fine\_tuning$      & 0.44 & 0.19 & 0.27 \\ \cline{2-5} \cline{7-10}
                                & $Longformer-Fine\_tuning$  & 0.53 & 0.25 & 0.37 &             & $Longformer-Fine\_tuning$   & 0.47 & 0.21 & 0.27 \\ \cline{2-5} \cline{7-10}
                                & $ProphetNet-Fine\_tuning$  & 0.54 & 0.25 & 0.37 &             & $ProphetNet-Fine\_tuning$   & 0.46 & 0.21 & 0.26 \\ \hline 

                                & $ATSumm$                   & \bf0.48 & \bf0.24 & \bf0.30 &    & $ATSumm$                   & \bf0.46 & \bf0.26 & \bf0.36 \\ \cline{2-5} \cline{7-10}
                                & $T5-Fine\_tuning$          & 0.45 & 0.17 & 0.27 &             & $T5-Fine\_tuning$           & 0.44 & 0.23 & 0.32 \\ \cline{2-5} \cline{7-10}
    ${D_3}$ & $BART-Fine\_tuning$        & 0.43 & 0.16 & 0.25 & ${D_{10}}$ & $BART-Fine\_tuning$ & 0.42 & 0.21 & 0.33 \\ \cline{2-5} \cline{7-10}
                                & $PEGASUS-Fine\_tuning$     & 0.44 & 0.17 & 0.26 &             & $PEGASUS-Fine\_tuning$      & 0.39 & 0.18 & 0.29 \\ \cline{2-5} \cline{7-10}
                                & $Longformer-Fine\_tuning$  & 0.45 & 0.19 & 0.28 &              & $Longformer-Fine\_tuning$  & 0.44 & 0.23 & 0.32 \\ \cline{2-5} \cline{7-10}
                                & $ProphetNet-Fine\_tuning$  & 0.45 & 0.20 & 0.28 &             & $ProphetNet-Fine\_tuning$   & 0.42 & 0.23 & 0.33 \\ \hline 

                                & $ATSumm$                   & \bf0.48 & \bf0.21 & \bf0.29 &    & $ATSumm$                   & \bf0.49 & \bf0.24 & \bf0.34 \\ \cline{2-5} \cline{7-10}
                                & $T5-Fine\_tuning$          & 0.45 & 0.17 & 0.26 &             & $T5-Fine\_tuning$           & 0.43 & 0.20 & 0.30 \\ \cline{2-5} \cline{7-10}
    ${D_7}$ & $BART-Fine\_tuning$        & 0.46 & 0.19 & 0.27 & ${D_{12}}$ & $BART-Fine\_tuning$ & 0.45 & 0.19 & 0.31 \\ \cline{2-5} \cline{7-10}
                                & $PEGASUS-Fine\_tuning$     & 0.45 & 0.17 & 0.27 &             & $PEGASUS-Fine\_tuning$      & 0.43 & 0.18 & 0.32 \\ \cline{2-5} \cline{7-10}
                                & $Longformer-Fine\_tuning$  & 0.46 & 0.19 & 0.27 &             & $Longformer-Fine\_tuning$   & 0.46 & 0.20 & 0.31 \\ \cline{2-5} \cline{7-10}
                                & $ProphetNet-Fine\_tuning$  & 0.44 & 0.18 & 0.26 &             & $ProphetNet-Fine\_tuning$   & 0.44 & 0.19 & 0.29 \\ \hline   
    \end{tabular}} 
\end{table*}

\subsection{Comparison with Fine-tuned Transformer models}\label{s:finetune}
\par In this Subsection, we compare and validate the effectiveness of ATSumm with the existing state-of-the-art transformer models fined-tuned on our proposed ARIES dataset. For this, we consider $5$ different transformer models, i.e., T5~\cite{raffel2020exploring}, BART~\cite{lewis2020bart}, PEGASUS~\cite{zhang2020pegasus}, Longformer~\cite{beltagy2020longformer}, and ProphNet~\cite{qi2020prophetnet}. We then fine-tune these transformer models on the ARIES dataset using the following parameters: batch size is $16$, maximum output summary length is $200$ tokens, and others are default parameters. We decide these parameter values using an experiment on randomly selected two datasets by varying the batch size value from $4$ to $32$ for fine-tuning. We found the batch size $16$ gives high efficiency compared to other batch sizes and then chose this value for this experiment. For our experiment, we randomly select six different disaster datasets: $D_1$, $D_3$, $D_7$, $D_8$, $D_{10}$, and $D_{12}$. Our results, as shown in Table~\ref{table:finetune}, indicate that the ATSumm outperforms with the fine-tuned transformer models by $3.92-17.24$\%, $9.52-33.33$\%, and $5.88-19.51$\% in terms of ROUGE-1, ROUGE-2, and ROUGE-L F1-score, respectively. Furthermore, we observe that the performance increase of ATSumm is highest when compared to the fine-tuned PEGASUS model and lowest when compared to the fine-tuned Longformer model. 
Therefore, based on the above experiment, we conclude that ATSumm ensures high performance with the existing transformer models even after being fined-tuned on disaster events.

\subsection{Human Evaluation}\label{s:humeval}
\par As the abstractive summary is more closer to the human written summaries, in this Subsection, we perform a human evaluation to evaluate the quality of the summary generated by ATSumm. Following~\cite{vadlapudi2010automated, iskender2021reliability, fabbri2021summeval}, we compare the summary generated by ATSumm with the baseline summaries based on five different quality metrics:

\begin{itemize}
    \item \textbf{Fluency-} The summary must follow the grammatical rules without any formatting problems in order to read well to the readers. 
    
    \item \textbf{Readability-} The summary should be easy to read and understandable to the readers.  
    
    \item \textbf{Conciseness-} The summary should contain no unnecessary words.   
    
    \item \textbf{Relevance-} The summary should only contain relevant information from the given input tweets. 

    \item \textbf{Non-redundancy-} The summary must be unique, with no two tweets containing the same information in the summary. 
\end{itemize}

\begin{table*}[!ht]
    \centering 
    \caption{Table shows the Aggregated (average) Fluency Score (\textbf{AFS}), Readability Score (\textbf{AReS}), Conciseness Score (\textbf{ACS}), Relevance Score (\textbf{ARS}), and Non-redundancy Score (\textbf{ANS}) of the summary generated by ATSumm and different baselines on four randomly selected datasets: $D_1$, $D_4$, $D_9$, and $D_{12}$.}
    \label{table:human}
    \resizebox{0.85\linewidth}{!}{\begin{tabular}{|c|l|c|c|c|c|c|c|l|c|c|c|c|c|} \hline
        \textbf{Dataset} & \textbf{Approach} & \textbf{AFS} & \textbf{AReS} & \textbf{ACS} & \textbf{ARS} & \textbf{ANS}& \textbf{Dataset} & \textbf{Approach} & \textbf{AFS} & \textbf{AReS} & \textbf{ACS} & \textbf{ARS} & \textbf{ANS} \\ \hline
        

                            & $ATSumm$      & \bf4.25 & \bf4.59 & \bf4.62 & \bf4.42 & \bf4.33 &     & $ATSumm$     & \bf4.61 & \bf4.75 & \bf4.50 & \bf4.75 & \bf4.69 \\ \cline{2-7} \cline{9-14}
                            & $CowABS$   & 3.08 & 3.41 & 3.52 & 3.33 & 3.58 &                    & $CowABS$  & 3.41 & 3.50 & 3.25 & 3.33 & 3.50 \\ \cline{2-7} \cline{9-14}
                            & $PGN$         & 2.75 & 2.82 & 2.91 & 2.66 & 2.91 &                    & $PGN$        & 3.16 & 2.98 & 2.80 & 3.08 & 3.33 \\ \cline{2-7} \cline{9-14}
                            & $PTESumm$     & 2.91 & 3.08 & 3.12 & 2.96 & 3.00 &                    & $PTESumm$    & 3.69 & 3.56 & 3.72 & 3.50 & 3.75 \\ \cline{2-7} \cline{9-14}
                            & $PSumm$       & 3.33 & 3.75 & 3.85 & 3.91 & 3.85 &                    & $PSumm$      & 3.92 & 3.96 & 3.83 & 3.91 & 3.50 \\ \cline{2-7} \cline{9-14}
                            & $CLiQS\_CM$   & 3.98 & 4.05 & 3.83 & 4.00 & 3.83 &                    & $CLiQS\_CM$  & 4.05 & 4.16 & 4.00 & 4.25 & 4.25 \\ \cline{2-7} \cline{9-14}
                            & $T5$          & 4.05 & 4.16 & 4.00 & 4.25 & 4.16 &                    & $T5$         & 4.25 & 4.50 & 4.25 & 4.25 & 4.50 \\ \cline{2-7} \cline{9-14}
                    ${D_1}$ & $BART$        & 3.85 & 3.69 & 3.91 & 3.83 & 3.75 &            ${D_9}$ & $BART$  & 4.14 & 4.25 & 4.25 & 4.16 & 4.32 \\ \cline{2-7} \cline{9-14}
                            & $PEGASUS$     & 3.91 & 3.83 & 3.75 & 3.66 & 3.83 &                    & $PEGASUS$    & 4.14 & 4.00 & 4.05 & 4.00 & 4.16 \\ \cline{2-7} \cline{9-14}
                            & $Longformer$  & 4.00 & 4.25 & 4.05 & 4.16 & 3.75 &                    & $Longformer$ & 3.91 & 4.00 & 3.83 & 3.83 & 4.05 \\ \cline{2-7} \cline{9-14}
                            & $ProphetNet$  & 3.91 & 4.00 & 3.83 & 4.16 & 3.91 &                    & $ProphetNet$ & 3.83 & 3.98 & 3.91 & 3.91 & 4.00 \\ \hline

                            & $ATSumm$      & \bf4.41 & \bf4.50 & \bf4.75 & \bf4.65 & \bf4.50 &   & $ATSumm$     & \bf4.75 & \bf4.69 & \bf4.75 & \bf4.50 & \bf4.61 \\ \cline{2-7} \cline{9-14}
                            & $CowABS$   & 2.91 & 2.66 & 2.82 & 2.82 & 2.75 &     & $CowABS$  & 3.58 & 3.50 & 3.33 & 3.41 & 3.66 \\ \cline{2-7} \cline{9-14}
                            & $PGN$         & 2.83 & 2.66 & 2.82 & 2.75 & 2.91 &     & $PGN$        & 3.33 & 3.12 & 3.41 & 3.41 & 3.58 \\ \cline{2-7} \cline{9-14}
                            & $PTESumm$     & 3.03 & 2.96 & 2.91 & 3.00 & 3.12 &     & $PTESumm$    & 3.56 & 3.75 & 3.50 & 3.96 & 3.90 \\ \cline{2-7} \cline{9-14}
                            & $PSumm$       & 3.69 & 3.75 & 3.56 & 3.42 & 3.12 &     & $PSumm$      & 4.25 & 4.25 & 4.41 & 4.25 & 4.16 \\ \cline{2-7} \cline{9-14}
                            & $CLiQS\_CM$   & 3.91 & 4.05 & 4.14 & 4.05 & 4.25 &     & $CLiQS\_CM$  & 4.00 & 4.14 & 4.05 & 3.98 & 4.00 \\ \cline{2-7} \cline{9-14}
                            & $T5$          & 4.05 & 4.16 & 4.25 & 4.00 & 4.14 &     & $T5$         & 3.91 & 4.00 & 4.16 & 4.00 & 4.05 \\ \cline{2-7} \cline{9-14}
                    ${D_4}$ & $BART$        & 3.91 & 3.99 & 4.00 & 3.83 & 3.75 & ${D_{12}}$ & $BART$  & 3.96 & 4.00 & 4.12 & 3.92 & 3.96 \\ \cline{2-7} \cline{9-14}
                            & $PEGASUS$     & 4.25 & 4.25 & 4.32 & 4.25 & 4.36 &     & $PEGASUS$    & 4.00 & 3.98 & 4.05 & 3.91 & 3.91 \\ \cline{2-7} \cline{9-14}
                            & $Longformer$  & 4.05 & 4.00 & 3.99 & 4.16 & 3.83 &     & $Longformer$ & 3.98 & 4.00 & 3.91 & 3.83 & 3.98 \\ \cline{2-7} \cline{9-14}
                            & $ProphetNet$  & 4.14 & 3.83 & 4.00 & 4.05 & 3.91 &     & $ProphetNet$ & 4.05 & 3.98 & 4.00 & 4.14 & 4.05 \\ \hline 
    \end{tabular} }
\end{table*}

We employ three meta-annotators\footnote{We select these meta-annotators through the Quality Assessment Evaluation proposed by \cite{garg2023portrait}.} for evaluating the summaries, taking care not to select the same annotators who generated the summaries. The meta-annotators are post-graduate students between the age group of $25-30$ who speak English fluently. We follow existing research works~\cite{poddar2022caves, garg2023portrait}, where each annotator individually reads the summaries generated by ATSumm and the different baselines and then assigns a score to each of the summaries in the range of $1$ (lowest score) - $10$ (highest score) based on metrics defined above. We get all of the above-mentioned scores from all three meta-annotators for each summary. To merge scores from different meta-annotators, we aggregate all of them and average them. For example, if the Fluency Score of a summary from three meta-annotators is x, y, and z, then the Aggregate Fluency Score is (x+y+z)/3. Similarly, we calculated the Aggregate Readability Score, Aggregate Conciseness Score, Aggregate Relevance Score and Aggregate Non-redundancy Score. We show the aggregated score over all three meta-annotators for four randomly selected datasets: $D_1$, $D_4$, $D_9$, and $D_{12}$ in Table~\ref{table:human}. Our observations indicate that the quality of the ATSumm summary is better than existing summarization approaches by $3.11-40.88$\%. We also observe that the aggregated score of ATSumm is above $4.25$ for all five metrics across the datasets. 
The above results show that the summaries generated by ATSumm are of higher quality than the other baseline approaches. 

\subsection{Human A/B Testing} \label{s:humAB}
\par In this Subsection, we design a human A/B testing~\cite{iskender2021reliability, goyal2022news} to ensure the effectiveness of our proposed model against the existing state-of-the-art models. In this experiment, we consider two models, A and B, where A could be our model (ATSumm), and B could be the considered baselines (discussed in Subsection~\ref{s:base}) or the other way round. In any case, the annotators will not know which model is A or B. We employ three annotators who are students in the age group of $20-30$ and take care not to select the same annotators who generated the summaries. Each annotator goes through all the summaries generated by all these models of a dataset and picks the best summary from them. An annotator may select a Tie if he/she finds the summary of both models containing the same information. In the case of Tie, we bring in a $4^{th}$ annotator and perform the same experiment. We repeat this procedure for four datasets: $D_1$, $D_4$, $D_9$, and $D_{12}$, and each dataset is then passed to various models for recording the response of each model. Once we have the ratings for each model’s comparison with ours, we maintain records of how often our model wins, loses, or ties with the other model out of the four datasets. Our results, as shown in Table~\ref{table:humanAB}, indicate that the summary generated by ATSumm has a higher win percentage than the loss or tie percentage, showing how well-appreciated our model is. Therefore, based on the above experiment, we can say that the generated summary quality by ATSumm is high and is highly preferred over the other various models.

\begin{table}[!ht]
    \centering 
    \caption{Table shows the results of human A/B testing for assessing the summary generated by ATSumm and different baseline models. We show the average win, loss, and tie percentage of three annotators scores on four different disaster datasets.}
    \label{table:humanAB}
    \resizebox{0.65\linewidth}{!}{\begin{tabular}{|l|c|c|c|} \hline
        \textbf{Model} & \multicolumn{3}{c|}{\textbf{ATSumm Vs Model}} \\ \cline{2-4}
        
        \textbf{} & \textbf{Win\%} & \textbf{Loss\%} & \textbf{Tie\%} \\ \hline

        $CowABS$   & 100   & -     & -     \\ \hline
        $PGN$         & 100   & -     & -     \\ \hline
        $PTESumm$     & 91.67 & -     & 8.33  \\ \hline
        $PSumm$       & 75.00 & 8.33  & 16.67 \\ \hline
        $CLiQS\_CM$   & 83.34 & 8.33  & 8.33  \\ \hline
        $T5$          & 66.66 & 16.67 & 16.67 \\ \hline
        $BART$        & 75.00 & 16.67 & 8.33  \\ \hline
        $PEGASUS$     & 75.00 & -     & 25.00 \\ \hline
        $Longformer$  & 91.67 & 8.33  & -     \\ \hline
        $ProphetNet$  & 83.34 & -     & 16.67 \\ \hline
    \end{tabular} }
\end{table}

\section{Conclusions and Future works} \label{s:con}
\par In this paper, we propose a deep learning-based framework ATSumm, which generates a human-readable summary of a disaster with sparse training data. The main challenge of any deep learning-based approach is that it requires huge training data. The sparsity of the training data is handled using domain knowledge in the form of auxiliary information. The first phase of ATSumm (extractive phase) ranks the tweets using domain knowledge from an existing ontology; while we provide key-pharse and their importance score as auxiliary information in the second phase of ATSumm (abstractive phase). To the best of our knowledge, our proposed ATSumm is the first model on abstractive disaster tweet summarization, which utilizes auxiliary information in a deep-learning framework to handle the problem of sparse training data. Both quantitative and qualitative experiments confirm the superiority of ATSumm over existing state-of-the-art models. Our quantitative experimental analysis shows that ATSumm can ensure $4-80\%$ higher ROUGE-N F1-score than existing state-of-the-art baseline approaches. Additionally, our qualitative experiments analysis shows that ATSumm can ensure $3.11-40.88\%$ better performance than existing summarization approaches for all five qualitative metrics across the datasets. Apart from ATSumm, in this paper, we generate and publicly provide a novel ARIES dataset consisting of $7500$ training samples and $13$ different disaster events with corresponding ground-truth summaries. As future work, to handle the data sparsity problem, we can look further into more auxiliary information, which we can utilize along with key-phrases and their importance scores. Additionally, we wish to apply a similar model for analyzing domains where such data sparsity is generic, like insurance fraud detection, patient record summarization, etc.

\bibliographystyle{elsarticle-num}
\bibliography{bibtext}

\end{document}